\documentclass{article}

\PassOptionsToPackage{numbers,sort&compress}{natbib}
\usepackage[preprint]{neurips_2025}

\usepackage[utf8]{inputenc} 
\usepackage[T1]{fontenc}    
\usepackage{hyperref}       
\usepackage{url}            
\usepackage{booktabs}       
\usepackage{amsfonts}       
\usepackage{nicefrac}       
\usepackage{microtype}      
\usepackage{wrapfig}
\usepackage{graphicx}
\usepackage{enumitem}
\usepackage{amsmath}
\usepackage{multirow}
\usepackage{rotating}
\usepackage{booktabs,pifont}
\usepackage[ruled,vlined,linesnumbered]{algorithm2e}

\usepackage{amssymb}
\usepackage{marvosym}

\usepackage[table,dvipsnames]{xcolor}

\newcommand{\cmark}{\textcolor{ForestGreen}{\ding{51}}}
\newcommand{\xmark}{\textcolor{red}{\ding{55}}}

\title{MathSticks: A Benchmark for Visual Symbolic Compositional Reasoning with Matchstick Puzzles}

\author{
\small Yuheng Ji$^{1,2,3,*}$, Huajie Tan$^{3,4,*}$, Cheng Chi$^{3,*}$, Yijie Xu$^{3,5}$, Yuting Zhao$^{1,2}$, Enshen Zhou$^{3,6}$, \\
\small Huaihai Lyu$^{1,2,3}$, Pengwei Wang$^{3}$, Zhongyuan Wang$^{3}$, Shanghang Zhang$^{3,4,\text{\Letter}}$, Xiaolong Zheng$^{1,2,\text{\Letter}}$ \\
$^1$ \small Institute of Automation, Chinese Academy of Sciences
$^2$ \small School of Artificial Intelligence, \\University of Chinese Academy of Sciences
$^3$ \small Beijing Academy of Artificial Intelligence \\
$^4$ \small Peking University
$^5$ \small The University of Sydney
$^6$ \small Beihang University
}

\begin{document}

\maketitle

\let\thefootnote\relax\footnotetext{$^{*}$ Equal contribution. $^{\text{\Letter}}$ Corresponding author.}

\begin{abstract}

We introduce \textsc{MathSticks}, a benchmark for Visual Symbolic Compositional Reasoning (VSCR), which unifies visual perception, symbolic manipulation, and arithmetic consistency. Each task presents an incorrect matchstick equation that must be corrected by moving one or two sticks under strict conservation rules. 
The benchmark includes both text-guided and purely visual settings, systematically covering digit scale, move complexity, solution multiplicity, and operator variation, with 1.4M generated instances and a curated test set.
Evaluations of 14 vision–language models reveal substantial limitations: closed-source models succeed only on simple cases, open-source models fail in the visual regime, while humans exceed 90\% accuracy. 
These findings establish \textsc{MathSticks} as a rigorous testbed for advancing compositional reasoning across vision and symbols. Our code and dataset are publicly available at \url{https://github.com/Yuheng2000/MathSticks}.

\end{abstract}
    
\section{Introduction}
\label{sec:intro}

Existing tasks in vision and reasoning often tackle only one dimension, such as visual perception, reasoning, or arithmetic computation, without forming a unified challenge.
To bridge this gap, we introduce Visual Symbolic Compositional Reasoning (VSCR), which unifies all three by linking visual perception with symbolic manipulation.
VSCR requires models to recognize structured elements, plan transformations under explicit constraints, and verify arithmetic consistency, reflecting core aspects of everyday cognition.
Yet current vision–language models (VLMs) struggle with symbol-level understanding and constrained edits, while existing benchmarks remain limited. As summarized in Tab.~\ref{tab:task_setting_comparison}, they lack:
\emph{(i)} constrained symbolic manipulation with solvability guarantees;
\emph{(ii)} fine-grained control of task settings and difficulty; and
\emph{(iii)} evaluation protocols isolating pure-visual reasoning.

\begin{wrapfigure}{r}{0.25\linewidth}
    \centering
    \vspace{-1em}
    \includegraphics[width=1.0\linewidth]{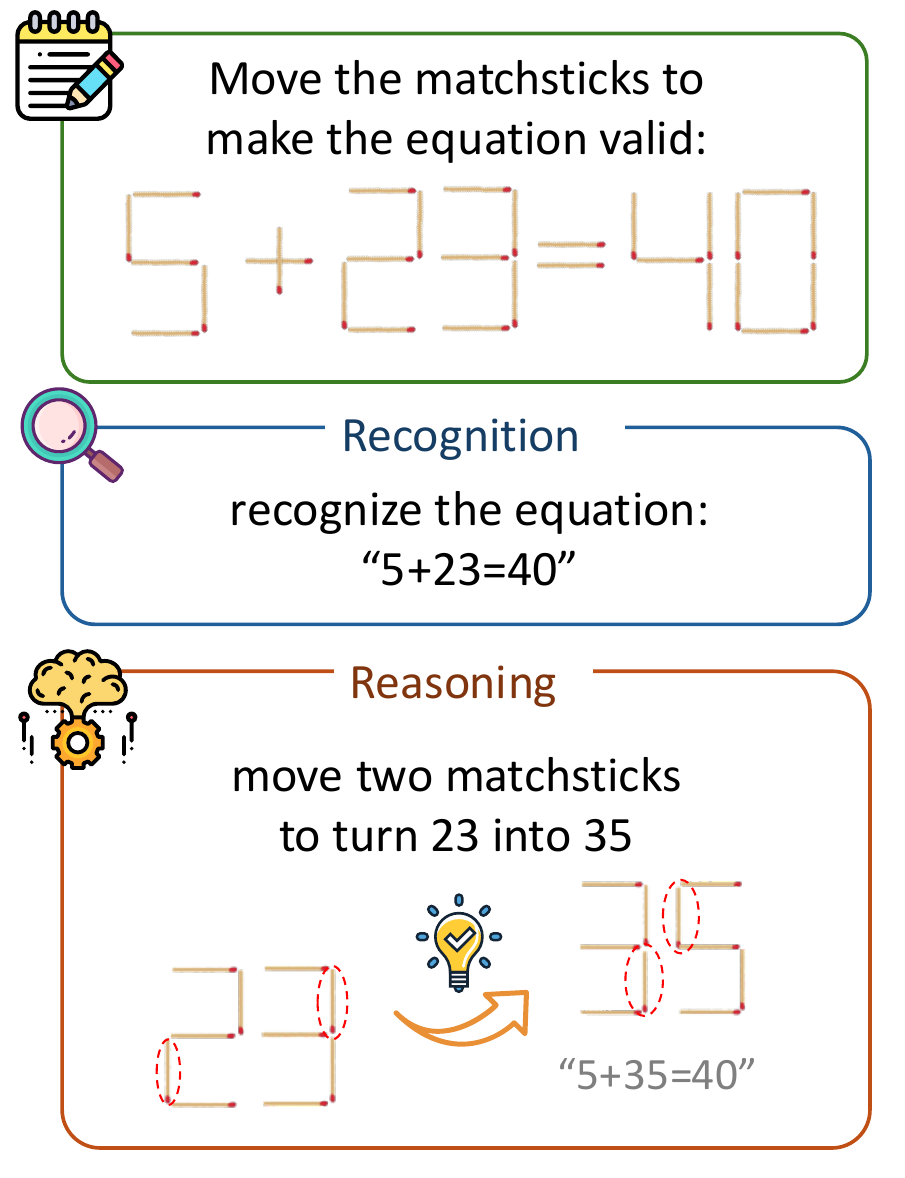}
    \vspace{-1.5em}
    \caption{Overview of the \textsc{MathSticks} task.}
    \vspace{-4em}
    \label{fig:overview}
\end{wrapfigure}

To close these gaps, we present \textsc{MathSticks}, a benchmark built on matchstick arithmetic puzzles.
Each task is an incorrect equation corrected by moving one or two sticks, enforcing both stick conservation and arithmetic consistency.
Two regimes are defined: a text-prompted setting, where the equation string is provided, and a pure-visual setting, where only the rendered puzzle is shown (Fig.~\ref{fig:overview}).
This separation enables diagnosis of symbolic reasoning versus visual parsing. The pipeline systematically enumerates digit-level operations, validates solvability, and generates 1.4M instances, from which a balanced evaluation set of 400 items is released.
The benchmark spans digit scale (Levels~1–4), move complexity, solution multiplicity, and operator flipping, while avoiding extreme cases to ensure human accessibility.

We evaluate 14 representative VLMs, including closed-source models (e.g., o3, Gemini~2.5~Pro) and open-source families (Qwen2.5-VL, InternVL).
Results show a clear capability gap: closed models solve simple puzzles but falter on multi-move and operator-flip cases, open-source models score near zero in the pure-visual regime, and humans consistently exceed 90\% accuracy. This establishes \textsc{MathSticks} as a controlled, diagnostic testbed for advancing VSCR.

\begin{table*}[t]
\small
\centering
\caption{Comparison of representative benchmarks across five key dimensions. \textit{Partial} denotes limited coverage in specific sub-tasks or without a unified protocol.}
\vspace{0.5em}
\scalebox{0.9}{
\begin{tabular}{lccccc}
\toprule
\textbf{Benchmark} & \textbf{Valid editing} & \textbf{Executable check} & \textbf{Difficulty slices} & \textbf{Symbolic grounding} & \textbf{Scale ($\approx$1M)} \\
\midrule
CLEVR~\cite{johnson2017clevr}              & \xmark & \cmark & \xmark & \textit{partial} & \xmark \\
CLEVR-Change~\cite{park2019robust}         & \xmark & \xmark & \xmark & \xmark           & \xmark \\
MathVista~\cite{lumathvista}               & \xmark & \textit{partial} & \xmark & \textit{partial} & \xmark \\
AI2D~\cite{hiippala2021ai2d}               & \xmark & \xmark & \xmark & \textit{partial} & \xmark \\
ChartQA~\cite{masry2022chartqa}            & \xmark & \textit{partial} & \xmark & \textit{partial} & \xmark \\
TRANCE~\cite{hong2021transformation}       & \cmark & \xmark & \textit{partial} & \textit{partial} & \xmark \\
VisualTrans~\cite{ji2025visualtrans}       & \cmark & \xmark & \textit{partial} & \xmark           & \xmark \\
\textbf{MathSticks (Ours)}                 & \cmark & \cmark & \cmark & \cmark           & \cmark \\
\bottomrule
\end{tabular}
}
\vspace{-1em}
\label{tab:task_setting_comparison}
\end{table*}

\section{MathSticks}

\textbf{Task Definition.}  
Each puzzle is an equation \(\textstyle x = [a, \oplus, b, =, c], \oplus \in \{+, -\}\),
rendered in matchstick-based seven-segment digits.  
The input is always an \emph{invalid} equation.  
The objective is to relocate $k \in \{1,2\}$ sticks, without insertion or deletion, such that the corrected form satisfies $a \oplus b = c$.  
This formulation integrates three components: visual perception of digits and operators, planning of valid symbolic transformations under strict constraints, and verification of arithmetic correctness.

For precise evaluation, every matchstick position is indexed at the segment level 
(e.g., \texttt{A0--A6}, \texttt{B0--B6}).  
During evaluation, VLMs are required to output edits in a canonical format such as 
\({\texttt{Move(A0, C3)}}\) for single moves or 
\({\texttt{Move(A0, C3), Move(E1, F4)}}\) for two moves.  
This ensures that predictions are machine-parsable, unambiguous, and directly verifiable against ground-truth solutions. Detailed visual conventions are provided in Appendix~\ref{app:visual_representation}.

\textbf{Categorization.}  
To support fine-grained diagnosis, we annotate each instance along four axes:  
(i) digit scale (Levels 1–4, from single- to two-digit operands),  
(ii) move complexity (1- vs.\ 2-move solvability),  
(iii) solution multiplicity (unique vs.\ multiple), and  
(iv) operator flipping (sign-changed vs.\ preserved).  
These factors yield seven diagnostic categories, with precise definitions and illustrations deferred to the Appendix~\ref{app:fine-grained}.

\textbf{Dataset Construction.}  
We design a two-stage generation pipeline that guarantees both completeness in the symbolic search space and fidelity in the rendered visual stimuli.  
(1) \emph{Symbolic enumeration}: each equation is represented as a 7-slot state and exhaustively explored under legal one- and two-stick moves with stick conservation.  
Candidate edits are filtered by arithmetic validation, deduplicated across move types, and labeled with diagnostic attributes (digit scale, move complexity, solution multiplicity, operator flipping).  
(2) \emph{Visual rendering}: each symbolic state is deterministically mapped to an image via a template library of seven-segment digits and operator slots, ensuring one-to-one correspondence between symbolic edits and visible stick relocations.  
This construction yields $\sim$1.4M solvable instances with large-scale coverage and structured metadata.  
Algorithmic details and pseudocode are provided in Appendix~\ref{app:construction}.

\textbf{Dataset Statistics.}  
The benchmark comprises 1,411,388 validated solvable instances.  
Difficulty is highly skewed toward Level~4 (79.07\%), with Levels~1–3 contributing 0.11\%, 1.31\%, and 19.51\%, respectively.  
Most instances require two-stick edits (82.01\%), while one-stick solutions are rare (4.18\%); the remainder admit both one- and two-stick corrections (13.81\%).  
Regarding solution multiplicity, 43.12\% of instances have a unique correction, whereas 56.88\% admit multiple valid corrections.  
For reproducible evaluation, we additionally release a compact 400-item test set, stratified evenly across Levels~1–4 (100 per level).  
More detailed statistics are provided in Appendix~\ref{app:statistics}.

\vspace{-0.3em}
\section{Experiments}
\label{sec:experiments}

\subsection{Experimental Setup}
\label{sec:setup}

\textbf{Setup and Protocol.}
We evaluate models on the \textsc{MathSticks} benchmark under two input regimes:  
(i) \textit{text-prompted}, where the symbolic equation string is explicitly provided; and  
(ii) \textit{pure-visual}, where only the matchstick image is given, requiring OCR and structural parsing before subsequent reasoning.
Each instance is further categorized into four difficulty levels (L1–L4), 
ranging from equations composed entirely of single-digit operands to cases involving one or more two-digit numbers. 
This hierarchy captures progressively greater visual and structural complexity. 
Additional dimensions such as operation complexity, solution multiplicity, and operator flipping are included for comprehensive diagnosis, 
with detailed breakdowns deferred to Appendix~\ref{app:fine-grained}. The primary evaluation metric is accuracy (\%), computed per level and regime.

\textbf{Human Evaluation.}  
To provide a reference ceiling, we recruited three adult participants who independently solved the full benchmark under the \emph{pure-visual regime}, where only rendered matchstick equations were shown.  
Because digits and operators are visually unambiguous, the same accuracies apply to the text-prompted regime.  
On average, participants exceeded 90\% accuracy with about one minute per problem, confirming the solvability of all tasks and highlighting the gap to current models.  
Detailed per-level results and solution times are reported in Appendix~\ref{app:human_eval}.

\textbf{Evaluated Models.}
We include both closed and open-source VLMs.  
For \textit{closed models}, we evaluate o3~\cite{gpto3-o4-mini}, Gemini~2.5~Pro~\cite{comanici2025gemini25}, Gemini~2.5~Flash~\cite{comanici2025gemini25}, GPT-o4-mini~\cite{gpto3-o4-mini}, Seed-1.6-Thinking~\cite{seed2025seed1}, Seed~1.6~\cite{seed2025seed1}, Claude~Sonnet~4~\cite{Claude}, and GPT-4o~\cite{GPT-4o}.  
For \textit{open-source models}, we cover two representative families: Qwen2.5\mbox{-}VL (7B, 32B, 72B) ~\cite{bai2025qwen25vl}and InternVL3 (8B, 38B, 78B)~\cite{chen2024internvl25}.  
In addition to the aggregate results reported here, we provide a fine-grained breakdown across difficulty dimensions in Appendix~\ref{app:fine-grained}.

\begin{table*}[t]
    \scriptsize
    \centering
    \caption{\textbf{Results on the MathSticks Reasoning Benchmark.}
    We report accuracy (\%) across difficulty levels under two regimes (w/ and w/o text prompt). 
    The best results are in \textbf{bold} and the second-best are \underline{underlined}.
    \footnotesize{$^{\dagger}$ Human evaluation was conducted only under the pure-visual regime. 
    Since the rendered digits and operators are visually unambiguous, participants consistently recognized the equations without error; 
    thus, providing the equation string does not alter the task.}
    }
    \vspace{0.5em}
    \scalebox{0.95}{
    \begin{tabular}{l|ccccc|ccccc}
        \toprule
        \multirow{2}{*}{\textbf{Model}} & \multicolumn{5}{c|}{\textbf{MathSticks (w/ text prompt)}} & \multicolumn{5}{c}{\textbf{MathSticks (w/o text prompt)}} \\ 
        \cmidrule(lr){2-6} \cmidrule(lr){7-11}
        & L1 & L2 & L3 & L4 & AVG & L1 & L2 & L3 & L4 & AVG \\ 
        \midrule
        \rowcolor[HTML]{F2F2F2} \multicolumn{11}{l}{\textbf{Closed Models}} \\ \midrule
        o3-250416                & 73.00 & 56.00 & 56.00 & 55.00 & \textbf{60.00} & 69.00 & 37.00 & 30.00 & 18.00 & \textbf{38.50} \\
        Gemini-2.5-Pro-250506    & 67.00 & 41.00 & 35.00 & 38.00 & \underline{45.25} & 42.00 & 19.00 & 21.00 &  8.00 & \underline{22.50} \\
        Gemini-2.5-Flash-250520  & 53.00 & 19.00 & 19.00 & 15.00 & 26.50 & 22.00 &  7.00 &  5.00 &  0.00 &  8.50 \\
        GPT-o4-mini-250416       & 47.00 & 19.00 & 21.00 & 12.00 & 24.75 & 30.00 &  7.00 &  3.00 &  2.00 & 10.50 \\
        Seed-1.6-Thinking-250615 & 41.00 & 14.00 & 18.00 & 13.00 & 21.50 &  3.00 &  0.00 &  0.00 &  1.00 &  1.00 \\
        Seed-1.6-250615          &  8.00 &  1.00 &  1.00 &  3.00 &  3.25 &  2.00 &  0.00 &  0.00 &  0.00 &  0.50 \\
        Claude Sonnet 4          &  7.00 &  0.00 &  2.00 &  0.00 &  2.25 &  0.00 &  0.00 &  0.00 &  0.00 &  0.00 \\
        GPT-4o-241120            &  0.00 &  0.00 &  0.00 &  0.00 &  0.00 &  0.00 &  0.00 &  0.00 &  0.00 &  0.00 \\
        \midrule
        \rowcolor[HTML]{F2F2F2} \multicolumn{11}{l}{\textbf{Open-Source Models}} \\ \midrule
        Qwen2.5-VL-7B-Instruct   & 0.00 & 0.00 & 0.00 & 0.00 & 0.00 & 0.00 & 0.00 & 0.00 & 0.00 & 0.00 \\
        Qwen2.5-VL-32B-Instruct  & 0.00 & 0.00 & 0.00 & 0.00 & 0.00 & 0.00 & 0.00 & 0.00 & 0.00 & 0.00 \\
        Qwen2.5-VL-72B-Instruct  & 0.00 & 0.00 & 0.00 & 0.00 & 0.00 & 0.00 & 0.00 & 0.00 & 0.00 & 0.00 \\
        InternVL3-8B             & 0.00 & 0.00 & 0.00 & 0.00 & 0.00 & 0.00 & 0.00 & 0.00 & 0.00 & 0.00 \\
        InternVL3-38B            & 0.00 & 0.00 & 0.00 & 0.00 & 0.00 & 0.00 & 0.00 & 0.00 & 0.00 & 0.00 \\
        InternVL3-78B            & 0.00 & 0.00 & 0.00 & 0.00 & 0.00 & 0.00 & 0.00 & 0.00 & 0.00 & 0.00 \\
        \midrule
        \rowcolor[HTML]{F2F2F2} \multicolumn{11}{l}{\textbf{Human Performance}} \\ \midrule
        Human$^{\dagger}$       & 95.00 & 98.33 & 85.83 & 87.72 & 91.72 {\tiny $\pm$ 6.13} & 95.00 & 98.33 & 85.83 & 87.72 & 91.72 {\tiny $\pm$ 6.13} \\
        \bottomrule
    \end{tabular}
    }
    \label{tab:matchstick_results}
\end{table*}

\subsection{Performance Analysis}
The benchmark results in Tab.~\ref{tab:matchstick_results} reveal systematic differences across model families and input regimes, as well as between human and model performance.  
These comparisons provide a structured view of how current VLMs engage with symbolic constraints, and where the main performance gaps emerge.  

\textbf{Closed-source models substantially outperform open-source systems.}  
Proprietary models attain markedly higher accuracies, whereas open-source baselines remain at or near chance level across both regimes.  
The effect is large and consistent across difficulty levels, underscoring the challenge of structured reasoning for today’s open-source pipelines.  

\textbf{Reasoning-optimized variants outperform their standard counterparts.}  
Within the closed-source group, models advertised as reasoning-enhanced achieve higher accuracy: o3 reaches 60.00\% compared with 0.00\% for GPT-4o; Gemini-2.5-Pro achieves 45.25\% compared with 26.50\% for Gemini-2.5-Flash; Seed-1.6-Thinking reaches 21.50\% compared with 3.25\% for its base model.  
These observations point to the importance of stepwise and structured reasoning mechanisms for handling compositional edits.  

\textbf{o3 exhibits the strongest overall performance.}  
It attains the highest average accuracy (49.25\%) across regimes, followed by Gemini-2.5-Pro (33.88\%).  
Its “thinking with image” paradigm, which integrates visual parsing with multi-step symbolic reasoning, appears particularly effective for puzzles requiring both perceptual discrimination and arithmetic verification.  

\textbf{Explicit textual input yields consistent gains.}  
Providing the equation string improves accuracy for all models (e.g., o3 improves from 38.50\% without text to 60.00\% with text).  
Without text, models must additionally perform OCR-like recognition of digits and operators, making perception a salient bottleneck for end-to-end performance.  

\textbf{Humans maintain a clear advantage.}  
Participants achieve an average accuracy of 91.72\%, well above the best model’s 49.25\%.  
This confirms that the benchmark is readily solvable for humans, while exposing persistent limitations of current VLMs in integrating perception and reasoning.  

Overall, these results delineate a consistent pattern: while closed-source reasoning-enhanced models make tangible progress, a wide gap remains to human-level strategies, which involve reliable multi-stage reasoning that current systems cannot yet replicate.

\vspace{-0.7em}
\subsection{Error Analysis}
\label{sec:error}
We identify five primary error types observed in model outputs.  
Each reflects a distinct weakness in the perception–reasoning–output pipeline, illustrated with representative cases.  

\textbf{Perception errors.}  
Models occasionally misread digits or operators in the seven-segment display.  
For example, a common failure is confusing visually similar digits such as ``5'' and ``6,'' or overlooking the short vertical bar required to distinguish ``1'' from ``7.''  
These errors reveal insufficient robustness in low-level visual parsing when symbolic cues are absent.  

\textbf{Edit-planning errors.}  
Some solutions violate the basic rules of matchstick manipulation, such as proposing a move from a non-existent segment or implicitly adding an extra stick.  
These mistakes suggest that models often lack an internalized notion of structural constraints governing legal edits.  

\textbf{Arithmetic-verification errors.}  
In certain cases, models generate edits that produce an equation which is visually valid but numerically incorrect.  
For instance, transforming $9-3=5$ into $9-2=5$ satisfies the stick-move requirement but results in an arithmetically false statement.  
Such cases show that models may stop at surface plausibility without executing a full arithmetic check.  

\textbf{Operator-handling errors.}  
Another weakness arises in tasks requiring operator changes.  
For example, when converting $7+2=9$ into a valid form, some models attempt digit-level edits while leaving the operator unchanged, or misplace the segment needed to form a minus.  
Although less frequent than other error types, such mistakes directly undermine equation validity.  

\textbf{Output-format errors.}  
Finally, several models deviate from the required structured output format (e.g., \texttt{Move(A0, C3)}), instead producing free-form natural language such as “move the top stick from the 9 to make it a 3.”  
These outputs prevent automatic evaluation even when the underlying reasoning is partially correct.  

Taken together, these error types align closely with the difficulty dimensions summarized in Appendix~\ref{app:fine-grained}—for instance, edit-planning errors are prevalent in two-stick puzzles, operator-handling errors concentrate in flip cases, and perception errors intensify under pure-visual inputs.  
\section{Conclusion}
\label{mian-sec5}

We introduced \textsc{MathSticks}, a benchmark targeting visual symbolic compositional reasoning (VSCR) through matchstick arithmetic puzzles. The benchmark enforces solvability guarantees, provides stratified difficulty slices, and supports both text-prompted and pure-visual evaluation, enabling controlled and diagnostic assessment. Experiments with 14 VLMs reveal that even strong closed-source systems struggle with multi-step edits and operator flips, while open-source models collapse in the pure-visual regime. Human participants, by contrast, exceed 90\% accuracy. These results highlight a substantial capability gap and establish \textsc{MathSticks} as a compact yet challenging testbed for future progress in perception–symbol reasoning.

\section*{Acknowledgments}
This work was supported by the Natural Science Foundation of China under Grant Nos. 72225011, 72434005, and 62476011, as well as by the National Science and Technology Major Project (No. 2022ZD0117800).




\bibliographystyle{unsrt}
\bibliography{main}  

\newpage
\appendix
\section*{Appendix}

This appendix provides supplementary materials that complement the main text.  
It covers related literature, dataset representation and construction, extended statistics, and additional experimental results.  
The sections are organized as follows:

\begin{itemize}[left=1em]
    \item Sec.~\ref{app:related} reviews related work, situating \textsc{MathSticks} within research on advanced visual reasoning, symbolic compositionality, and diagram- or math-based benchmarks.  
    \item Sec.~\ref{app:visual_representation} introduces the visual representation of matchsticks, detailing the segment-level indexing scheme that underlies all symbolic edits.  
    \item Sec.~\ref{app:construction} describes the dataset construction pipeline, including symbolic enumeration, arithmetic filtering, and deterministic visual rendering.  
    \item Sec.~\ref{app:statistics} reports detailed dataset statistics with both tabular summaries and distribution visualizations.  
    \item Sec.~\ref{app:more_exp} presents extended experimental results, including fine-grained analyses and human evaluation.  
    \item Sec.~\ref{app:prompts} lists the exact prompts used for evaluation under both text-prompted and pure-visual regimes, enabling full reproducibility.  
    \item Sec.~\ref{app:case_studies} presents qualitative case studies that illustrate representative model successes and common failure patterns.  
\end{itemize}

\section{Related Work}
\label{app:related}
\textbf{Vision–Language Models for Advanced Visual Reasoning.}  
Early work in multimodal learning largely concentrated on perception-oriented objectives such as recognition, captioning, and retrieval, often framed around improving robustness and generalization across diverse inputs~\cite{ji2025enhancing,ji2023learning,hao2025msc,hao2025really,zhao2025fastrsr,wu2025evaluating,lyu2025multiple,yao2025bi, lin2024draw, wang2025embodiedocc}.
With the advent of vision–language models (VLMs)~\cite{chen2024rh20t,zhou2024minedreamer,zhou2024code,zhou2025roborefer,an2025agfsync,qin2024worldsimbench,qin2024mp5,lyu2025egoprompt,bai2025alleviating, li2023enhancing, li2023general, li2024lamp, li2024mulsmo, li2024mlip, 10448413, 10687563, 10813576,zhang2025beyond,liu2024segment, luo2024llm}, the research focus has progressively shifted toward higher-level reasoning beyond static recognition.  
Recent studies highlight the effectiveness of chain-of-thought prompting and structured exploration strategies for enhancing compositional reasoning \cite{wei2022chain,zhouleast}, while reinforcement-based alignment further improves multi-step logical consistency \cite{lightman2023let,tan2025reason}.  
Parallel efforts extend these paradigms into embodied manipulation, where VLMs are optimized for long-horizon action planning and reasoning in physical environments \cite{ji2025robobrain,team2025robobrain,tan2025roboos,song2025maniplvm, li2025manipdreamer3d,liu2024robomamba,liu2025hybridvla}.  
Together, these advances suggest a broader trajectory: visual reasoning with large models is evolving from perception-centric tasks toward dynamic, multi-step, and embodiment-aware problem solving.

\textbf{Visual Symbolic Compositional Reasoning (VSCR).}  
VSCR requires a model to identify symbolic elements in images, plan reachable local edits under explicit constraints, and verify symbolic consistency after editing. This capability extends the scope of traditional compositional visual reasoning, which typically answers queries on a fixed scene without legal editing or post hoc verification. Early benchmarks such as CLEVR established controlled templates for attributes, relations, and counting, and spurred research on modular architectures and program-supervised reasoning \cite{johnson2017clevr}. Neuro-symbolic approaches further introduced explicit executors for logical verification \cite{yi2018neural,mao2019neuro}, while visual programming frameworks leverage large language models to synthesize executable procedures from images, modularizing perception and symbolic manipulation \cite{gupta2023visual}. However, these paradigms usually assume gold programs or fixed structures and rarely require searching for legal, solvable edits in the visual space. In contrast, our setting instantiates VSCR with matchstick equations and enforces stick conservation, restricted move budgets, operator flipping when legal, and arithmetic correctness, including the distinction between unique and multiple solutions.  

\textbf{Visual Transformation and Diagram- and Math-based Reasoning.}
Benchmarks on transformation and structural reasoning provide useful comparators but address different objectives. CLEVR-Change targets change captioning between image pairs without requiring algebraic verification ~\cite{hiippala2021ai2drst,lu2024mathvista,cui2023vtt,ji2025visualtrans}. TRANCE emphasizes transformation-driven reasoning in synthetic worlds but abstracts away from fine-grained symbolic edits and correctness guarantees \cite{hong2021transformation}. VisualTrans studies real-world human–object interactions, focusing on functional consequences of local modifications rather than constrained symbolic correction \cite{ji2025visualtrans}.  
Datasets on diagrams and mathematics highlight complementary reasoning skills. AI2D and AI2D\_RST examine parsing and relational understanding of instructional diagrams \cite{kembhavi2016diagram,hiippala2021ai2drst}. MathVista aggregates diverse visual math problems to test language–math compositionality \cite{lu2023mathvista}, while ChartQA and ScienceQA extend evaluation to chart-based reasoning and multimodal science questions \cite{masry2022chartqa,lu2022scienceqa}. Although valuable, these benchmarks primarily measure image-to-text answering. They seldom require a model to enumerate legal edits in the visual space and to verify symbolic correctness after modification.  

\textsc{MathSticks} complements these efforts by operationalizing VSCR as an end-to-end pipeline that integrates symbol recognition, constrained transformation planning, and arithmetic verification. It provides unified rendering rules, controllable difficulty slices (digit scale, move complexity, solution multiplicity, operator flipping), and large-scale coverage, enabling fine-grained diagnosis of visual–symbolic reasoning beyond prior QA- or captioning-centric tasks.

\begin{figure}[t]
    \centering
    \includegraphics[width=0.95\linewidth]{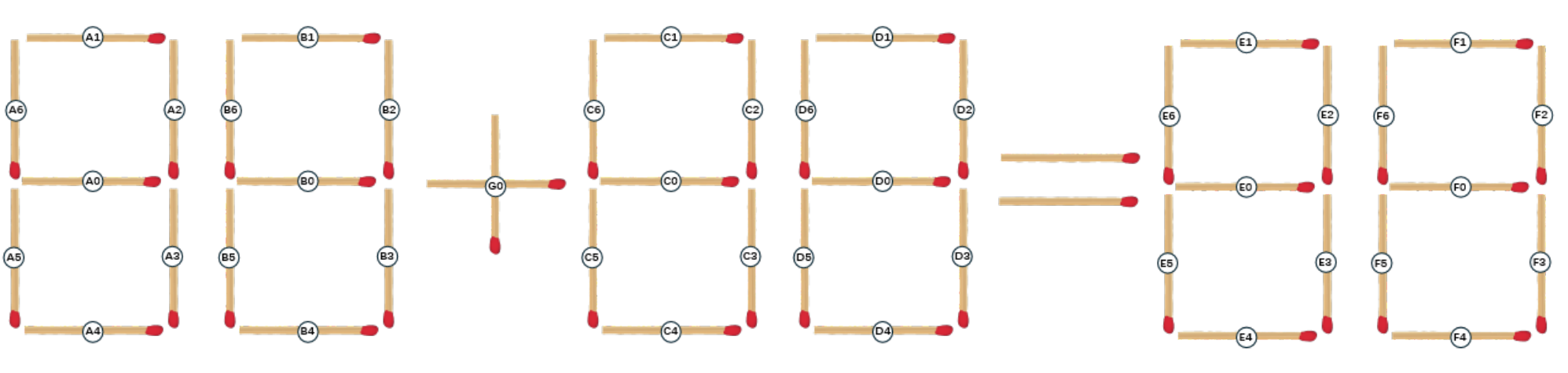}
    \caption{Figure 2: Illustration of the segment-level indexing scheme. Each digit position in the equation (indexed sequentially from left to right) is decomposed into seven labeled segments (0–6).}
    \label{fig:segment_labeling}
\end{figure}

\section{Visual Representation of Matchsticks}
\label{app:visual_representation}

To enable reproducible data generation and unambiguous parsing, we introduce a
\emph{segment-level indexing scheme} for all matchstick configurations.  
Each digit is rendered in a seven-segment layout, and every possible matchstick location is assigned a unique identifier, as shown in Fig.~\ref{fig:segment_labeling}.

\paragraph{Digit labeling.}  
For each operand and result digit, we adopt the standard seven-segment convention with indices
\(\{0,1,\dots,6\}\) referring to the horizontal and vertical strokes.  
As illustrated in Fig.~\ref{fig:segment_labeling}, segment indices are prefixed with a letter denoting the digit slot (A for the first digit, B for the second digit, etc.).  
For example:  
- \(\texttt{A0}\) denotes the bottom horizontal segment of digit A,  
- \(\texttt{A1}\) the top horizontal segment,  
- \(\texttt{A2}\) the top-right vertical segment,  
- \(\texttt{A3}\) the bottom-right vertical segment,  
- \(\texttt{A4}\) the bottom horizontal,  
- \(\texttt{A5}\) the bottom-left vertical,  
- \(\texttt{A6}\) the top-left vertical.  
The same scheme applies for all subsequent digit slots (B0–B6, C0–C6, etc.).

\paragraph{Operator labeling.}
The operator is placed in a dedicated slot \(G\).
We fix the horizontal bar as immutable, and define only one editable segment \texttt{G0} for the vertical stroke.
When \texttt{G0} is present, the operator is “\(+\)”; when absent, the operator is “\(-\)”.
Thus operator changes are realized by adding or removing the single stick at \texttt{G0}.
The equality sign “\(=\)” is only a visual separator, not indexed and not involved in edits.

\paragraph{Equation layout.}  
Each equation instance thus corresponds to a set of labeled matchsticks:
\[
\{ \texttt{A0--A6} \} \cup \{ \texttt{B0--B6} \} \cup \{ \texttt{C0--C6} \} \cup \{ \texttt{D0--D6} \} \cup \{ \texttt{E0--E6} \} \cup \{ \texttt{F0--F6} \}
\]
for digits, combined with operator segments (\(\texttt{G0}\)) and equality segments.  
This explicit labeling allows us to formalize puzzle states as discrete vectors,
supporting precise move operations such as  
\[
\boxed{\texttt{Move(A0, C3)}}
\]
which denotes relocating the stick from position \(\texttt{A0}\) to position \(\texttt{C3}\).

\paragraph{Usage in prompt construction.}  
This indexing scheme is also embedded in the evaluation prompt (see Sec.~\ref{app:prompts}), 
where models are required to output moves in the canonical format  
\(\boxed{\texttt{Move(<source>, <target>)}}\).  
The visual-to-symbolic mapping ensures that predictions are parsable, verifiable, and independent of rendering details.

\section{Dataset Construction Details}
\label{app:construction}

To ensure both completeness in the symbolic search space and fidelity in the visual representation, 
we propose a two-stage construction pipeline. 
First, \emph{symbolic enumeration} systematically explores all candidate equations and mines valid solutions under one- and two-stick moves. 
Second, \emph{visual rendering} deterministically assembles equation images from manually designed segment templates. 
This design guarantees large-scale coverage with precise alignment between symbolic transformations and visual stimuli.

\begin{algorithm}[!t]
\caption{Symbolic Enumeration and Solution Mining}
\label{alg:construction}
\DontPrintSemicolon
\SetKw{Continue}{continue}

\KwIn{Search ranges for $a,c,e\in\{-1,\ldots,9\}$ (tens slots), 
$b,d,f\in\{0,\ldots,9\}$ (units), and $g\in\{+,-\}$.}
\KwOut{A set $\mathcal{D}$ of solvable instances with diagnostic labels.}

\SetKwFunction{SoloToWhole}{SoloToWhole}
\SetKwFunction{EnumerateOne}{EnumerateOneStick}
\SetKwFunction{EnumerateTwo}{EnumerateTwoSticks}
\SetKwFunction{IsValid}{IsValidArithmetic}
\SetKwFunction{Labels}{AssignLabels}

$\mathcal{D}\leftarrow\varnothing$\;

\ForEach{$a,b,g,c,d,e,f$ in the Cartesian product}{
  $\mathbf{z}\leftarrow[a,b,g,c,d,e,f]$\;
  $(g^\ast,A,B,C)\leftarrow\SoloToWhole(\mathbf{z})$\;

  \If{\IsValid{$(g^\ast,A,B,C)$}}{
    \Continue\; \tcp{Original is already valid; skip as source}
  }

  $S_1 \leftarrow \EnumerateOne(\mathbf{z};\,\mathcal{T}_1)$\;
  $S_2 \leftarrow \EnumerateTwo(\mathbf{z};\,\mathcal{T}_2)$\;

  \tcp{Arithmetic filtering}
  $S_1^{\checkmark}\leftarrow \{\mathbf{z}'\in S_1:\IsValid(\SoloToWhole(\mathbf{z}'))\}$\;
  $S_2^{\checkmark}\leftarrow \{\mathbf{z}'\in S_2:\IsValid(\SoloToWhole(\mathbf{z}'))\}$\;

  \tcp{Deduplicate 2-stick solutions that also appear in 1-stick}
  $S_2^{\star}\leftarrow S_2^{\checkmark}\setminus S_1^{\checkmark}$\;

  \If{$|S_1^{\checkmark}|+|S_2^{\star}|>0$}{
    $\ell\leftarrow\Labels(\mathbf{z},S_1^{\checkmark},S_2^{\star})$\;
    $\mathcal{D}\leftarrow \mathcal{D}\cup\{(\mathbf{z},S_1^{\checkmark},S_2^{\star},\ell)\}$\;
  }
}
\KwRet{$\mathcal{D}$}\;
\end{algorithm}

\subsection{Symbolic Enumeration}

\paragraph{State encoding.}
Each equation instance is represented by a 7-slot tuple
\(
\mathbf{z}=[a,b,g,c,d,e,f]
\),
where \(a,b\) are the tens/units of the first operand,
\(c,d\) the tens/units of the second operand,
\(e,f\) the tens/units of the result,
and the operator slot \(g\in\{+,-\}\).
We allow blanks (coded as \(-1\)) for the tens slots \(a,c,e\).
A helper map \(\textsc{SoloToWhole}(\cdot)\) converts \(\mathbf{z}\) into
\((g,\, A,\, B,\, C)\) with
\(A=10\!\cdot\!\max(a,0)+b\),
\(B=10\!\cdot\!\max(c,0)+d\),
\(C=10\!\cdot\!\max(e,0)+f\).
Arithmetic validity is then checked directly:
the equation is valid if $A+B=C$ when $g{=}+$,
or if $A-B=C$ when $g{=}-$.

\paragraph{Move space.}
Seven-segment digits follow the standard 7-stick layout.
For each digit slot (A,B,C,\dots) and segment index \(s\in\{0,\ldots,6\}\),
a labeled stick position (e.g., A0, A1, …, A6) is defined (see Fig.~\ref{fig:segment_labeling}).
Two lookup tables encode legal edits:
\(\mathcal{T}_1\) for all \emph{single-stick} moves
and \(\mathcal{T}_2\) for all \emph{two-stick} composite edits
(including within-digit changes, cross-digit transfers, and operator flips),
all under stick conservation.

\paragraph{Operator labeling.}
The operator occupies a dedicated slot \(G\).
Its horizontal bar is fixed, while the vertical stroke is the only editable segment and is indexed as \texttt{G0}.
When \texttt{G0} is present the operator is “\(+\)”; when absent it is “\(-\)”.
Thus operator changes are realized by adding or removing the stick at \texttt{G0}.
The equality symbol “\(=\)” is only a visual separator in figures; it is not indexed and does not participate in edits.

\paragraph{Solution mining.}
For each \(\mathbf{z}\) we enumerate:
(i) all 1-stick reachable states \(S_1(\mathbf{z})\);
(ii) all 2-stick reachable states \(S_2(\mathbf{z})\).
After arithmetic filtering we deduplicate
\(S_2(\mathbf{z})\setminus S_1(\mathbf{z})\).
Each surviving transformation is labeled with
(digit scale: number of two-digit operands/results),
(move complexity: 1- vs.\ 2-stick),
(solution multiplicity),
and (operator flipping: whether \(+\leftrightarrow-\)).

The overall enumeration and filtering procedure is
summarized in Alg.~\ref{alg:construction}.

\subsection{Visual Rendering}

\begin{figure}[!t]
    \centering
    \includegraphics[width=\linewidth]{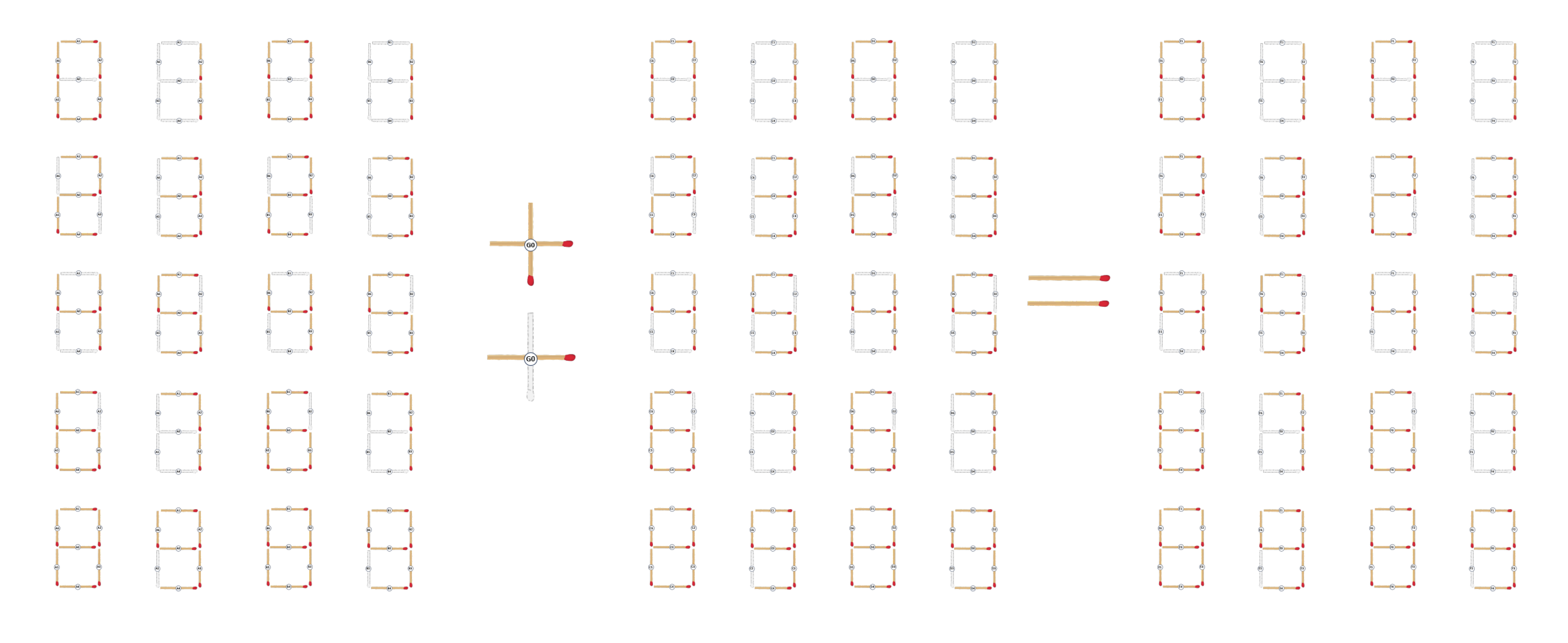}
    \caption{Example of the template library, showing digit slots with indexed 
    segments and the operator slot. Each index corresponds to a movable matchstick.}
    \label{fig:template_example}
\end{figure}

\paragraph{Template library.}
To standardize the visual construction, we manually created 
seven-segment digits (0–9) and the operator slot using a vector 
graphics editor.  
For each digit position (A, B, C, \(\ldots\)), we assigned unique indices 
to its seven possible segments, denoted as 
\(\texttt{A0}\ldots\texttt{A6}\), \(\texttt{B0}\ldots\texttt{B6}\), etc.,
ensuring one-to-one correspondence between indices and visible sticks.  
Each segment was drawn as an independent graphical object, and the operator 
slot was treated analogously (e.g., index \(\texttt{G0}\) for the horizontal bar).  
All assets were exported to PNG/SVG with fixed canvas size,
baseline alignment, and uniform spacing.  
An example of the template library is illustrated in Fig.~\ref{fig:template_example}.

\begin{algorithm}[!t]
\caption{Deterministic Visual Assembly}
\label{alg:render}
\DontPrintSemicolon
\KwIn{Symbolic equation $\mathbf{z}=[a,b,g,c,d,e,f]$;
template library $\mathcal{L}$ with per-slot segment-indexed assets.}
\KwOut{Rendered image $\mathsf{Img}(\mathbf{z})$.}

\SetKwFunction{Glyph}{FetchGlyph}
\SetKwFunction{Concat}{Concat}
\SetKwFunction{Layout}{Layout}

$\mathcal{G}\leftarrow []$ \tcp*{glyph list}
\ForEach{slot $s\in\{A,B,G,C,D,E,F\}$}{
  $v \leftarrow$ value in $\mathbf{z}$ at slot $s$\;
  $\mathcal{G}.\text{append}(\Glyph(\mathcal{L}, s, v))$\;
}
$\mathsf{Row}\leftarrow \Concat(\mathcal{G})$ \tcp*{fixed kerning and baseline}
$\mathsf{Img}(\mathbf{z})\leftarrow \Layout(\mathsf{Row})$\;
\KwRet{$\mathsf{Img}(\mathbf{z})$}\;
\end{algorithm}

\paragraph{Deterministic assembly.}
Given a symbolic equation \(\mathbf{z}=[a,b,g,c,d,e,f]\), rendering proceeds by slot-wise retrieval
and horizontal concatenation of the corresponding templates for slots \(\{A,B,G,C,D,E,F\}\).
Because indices are consistent across slots,
a symbolic edit such as \(\textsc{Move}(\texttt{A0},\texttt{C3})\)
corresponds to a unique visible stick relocation.
The full procedure is summarized in Alg.~\ref{alg:render}.

\paragraph{Reproducibility.}
All coordinates (origins, slot offsets, segment bounding boxes) are stored with the templates.
Thus any symbolic transformation applied to solvable instances \(\mathcal{D}\) 
(from the construction step in Sec.~\ref{alg:construction})
can be re-rendered without manual intervention,
ensuring consistent alignment between symbolic and visual forms.

\section{Detailed Dataset Statistics}
\label{app:statistics}

\begin{figure}[t]
    \centering
    \includegraphics[width=0.95\linewidth]{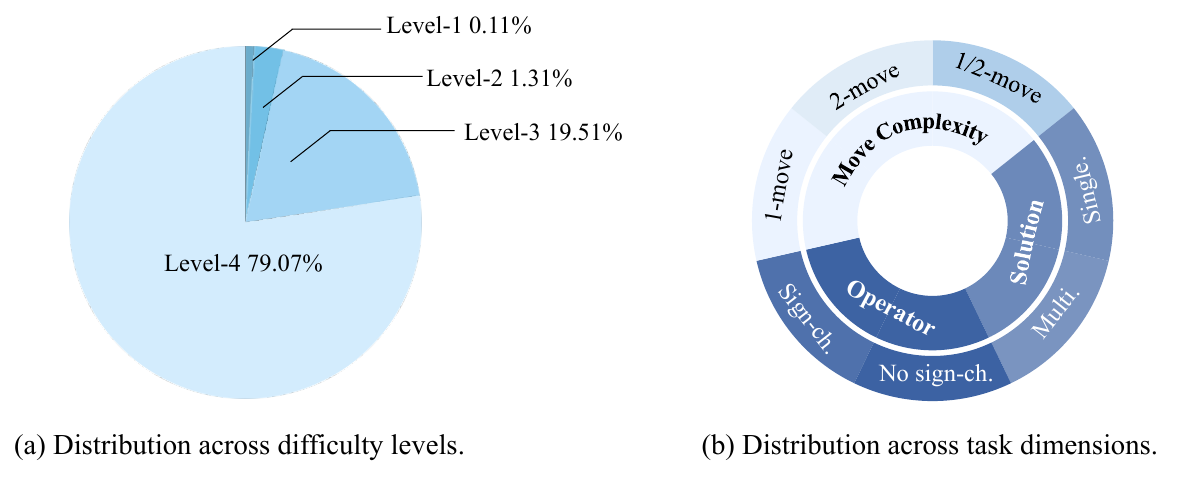}
    \caption{\textbf{Dataset distribution.} 
    (a) Proportions across difficulty levels. 
    (b) Decomposition by move complexity, solution multiplicity, and operator flipping.}
    \label{fig:dataset_distribution}
\end{figure}

This section provides a comprehensive breakdown of the \textsc{MathSticks} dataset, complementing the summary in the main text.  
In total, the benchmark contains 1,411,388 solvable instances.  
We analyze their distribution along three orthogonal axes: difficulty levels, move complexity, and solution multiplicity, as well as operator-flip requirements.  
Figure~\ref{fig:dataset_distribution} visualizes these dimensions, and Table~\ref{tab:dataset_statistics} reports detailed counts and percentages.  

\paragraph{Difficulty levels.}  
As shown in Fig.~\ref{fig:dataset_distribution}(a), Level~4 dominates the dataset (79.07\%), while Levels~1–3 contribute 0.11\%, 1.31\%, and 19.51\%, respectively.  
This skew reflects the combinatorial nature of more complex puzzles, where multi-digit configurations and multi-stick moves yield exponentially larger search spaces.  

\paragraph{Move complexity.}  
Most instances require two-stick transformations (82.01\%), with one-stick solutions accounting for only 4.18\%.  
The remaining 13.81\% can be solved by either a one-stick or two-stick move, illustrating the presence of multiple valid correction paths.  
This composition highlights the dataset’s emphasis on composite reasoning over local single-edit corrections.  

\paragraph{Solution multiplicity.}  
A substantial fraction of puzzles admit multiple valid edits.  
Across the dataset, 56.88\% of instances fall into this category, while 43.12\% have a unique correction.  
Multi-solution cases are particularly challenging for autoregressive models, which must converge on one valid output despite the ambiguity.  

\paragraph{Operator flipping.}  
Tasks involving operator changes (\(+ \leftrightarrow -\)) form a critical subspace, requiring models to edit abstract symbolic elements in addition to digit morphology.  
This dimension further stresses the need for integrating symbolic reasoning beyond perceptual transformation.  

\begin{table}[!t]
    \centering
    \caption{\textbf{Detailed dataset statistics.}
    Counts and percentages by difficulty level, further broken down by move complexity, solution multiplicity, and operator flipping.}
    \vspace{0.5em}
    \resizebox{\linewidth}{!}{
    \begin{tabular}{l|r r r|r r|r r}
        \toprule
        \multirow{2}{*}{\textbf{Level}} &
        \multicolumn{3}{c|}{\textbf{Move Complexity}} &
        \multicolumn{2}{c|}{\textbf{Solution Multiplicity}} &
        \multicolumn{2}{c}{\textbf{Operator}} \\
        \cmidrule(lr){2-4}\cmidrule(lr){5-6}\cmidrule(lr){7-8}
         & 1-move & 2-move & 1/2-move & Unique & Multiple & Flip & No flip \\
        \midrule
        L1 (1{,}505)      & 202 (13.42\%)   & 880 (58.47\%)   & 423 (28.11\%)   & 548 (36.41\%)   & 957 (63.59\%)   & 819 (54.42\%)  & 686 (45.58\%) \\
        L2 (18{,}466)     & 1{,}875 (10.15\%) & 14{,}340 (77.66\%) & 2{,}251 (12.19\%) & 11{,}692 (63.32\%) & 6{,}774 (36.68\%) & 6{,}743 (36.52\%) & 11{,}723 (63.48\%) \\
        L3 (275{,}406)    & 15{,}348 (5.57\%) & 219{,}715 (79.78\%) & 40{,}343 (14.65\%) & 127{,}208 (46.19\%) & 148{,}198 (53.81\%) & 105{,}185 (38.19\%) & 170{,}221 (61.81\%) \\
        L4 (1{,}116{,}011) & 41{,}505 (3.72\%) & 922{,}571 (82.67\%) & 151{,}935 (13.61\%) & 469{,}204 (42.04\%) & 646{,}807 (57.96\%) & 405{,}810 (36.36\%) & 710{,}201 (63.64\%) \\
        \midrule
        \rowcolor[HTML]{F2F2F2}
        Total (1{,}411{,}388) &
        58{,}930 (4.18\%) & 1{,}157{,}506 (82.01\%) & 194{,}952 (13.81\%) &
        608{,}652 (43.12\%) & 802{,}736 (56.88\%) &
        518{,}557 (36.74\%) & 892{,}831 (63.26\%) \\
        \bottomrule
    \end{tabular}}
    \label{tab:dataset_statistics}
\end{table}

\begin{table*}[!t]
    \scriptsize
    \centering
    \caption{\textbf{Results on the \textsc{MathSticks} Benchmark (w/ text prompt).}
    Accuracy (\%) is reported across categories reflecting move complexity, solution multiplicity, and operator change.
    The best performance is in \textbf{bold}, the second best is \underline{underlined}.}
    \vspace{0.5em}
    \scalebox{1.02}{
    \begin{tabular}{l|ccc|cc|cc}
        \toprule
        \multirow{2}{*}{\textbf{Model}} 
        & \multicolumn{3}{c|}{\textbf{Move Complexity}} 
        & \multicolumn{2}{c|}{\textbf{Solution Multiplicity}} 
        & \multicolumn{2}{c}{\textbf{Operator Change}} \\
        \cmidrule(lr){2-4} \cmidrule(lr){5-6} \cmidrule(lr){7-8}
        & \textbf{1-move} & \textbf{2-move} & \textbf{1/2-move} 
        & \textbf{Single-sol.} & \textbf{Multi-sol.} 
        & \textbf{Sign-changed} & \textbf{No sign-ch.} \\
        \midrule
        \rowcolor[HTML]{F2F2F2} \multicolumn{8}{l}{\textbf{Closed Models}} \\ \midrule
        o3-250416             & \textbf{82.50} & \textbf{52.88} & \textbf{89.18} & \textbf{47.79} & \textbf{71.69} & \textbf{51.45} & \textbf{65.19} \\
        Gemini-2.5-Pro-250506     & \underline{51.67} & \underline{38.43} & \underline{76.37} & \underline{28.93} & \underline{59.42} & \underline{38.91} & \underline{49.09} \\
        Gemini-2.5-Flash-250520   & 43.89 & 18.20 & 62.66 & 15.29 & 36.96 & 23.94 & 29.20 \\
        GPT-o4-mini-250416        & 59.17 & 14.71 & 66.19 & 15.60 & 33.05 & 21.15 & 26.95 \\
        Seed-1.6-Thinking-250615  & 31.39 & 15.17 & 41.72 & 10.89 & 30.06 & 16.83 & 24.80 \\
        Seed-1.6-250615           &  7.78 &  1.36 &  9.40 &  1.96 &  4.21 &  4.76 &  1.92 \\
        Claude Sonnet 4           &  7.78 &  0.78 &  5.63 &  1.21 &  2.81 &  2.85 &  1.57 \\
        GPT-4o-241120             &  0.00 &  0.00 &  0.00 &  0.00 &  0.00 &  0.00 &  0.00 \\
        \midrule
        \rowcolor[HTML]{F2F2F2} \multicolumn{8}{l}{\textbf{Open-Source Models}} \\ \midrule
        Qwen2.5-VL-7B-Instruct    & 0.00 & 0.00 & 0.00 & 0.00 & 0.00 & 0.00 & 0.00 \\
        Qwen2.5-VL-32B-Instruct   & 0.00 & 0.00 & 0.00 & 0.00 & 0.00 & 0.00 & 0.00 \\
        Qwen2.5-VL-72B-Instruct   & 0.00 & 0.00 & 0.00 & 0.00 & 0.00 & 0.00 & 0.00 \\
        InternVL3-8B              & 0.00 & 0.00 & 0.00 & 0.00 & 0.00 & 0.00 & 0.00 \\
        InternVL3-38B             & 0.00 & 0.00 & 0.00 & 0.00 & 0.00 & 0.00 & 0.00 \\
        InternVL3-78B             & 0.00 & 0.00 & 0.00 & 0.00 & 0.00 & 0.00 & 0.00 \\
        \bottomrule
    \end{tabular}
    }
    \label{tab:matchstick_results_textprompt}
\vspace{-1.0em}
\end{table*}

\section{More Experiments}
\label{app:more_exp}

\subsection{Fine-Grained Analysis}
\label{app:fine-grained}

To further dissect model behavior, we evaluate performance across fine-grained categories that capture different aspects of reasoning complexity. 
Specifically, we consider three orthogonal dimensions: \emph{move complexity} (one-stick vs.\ two-stick vs.\ mixed), 
\emph{solution multiplicity} (single-solution vs.\ multiple-solution instances), 
and \emph{operator flipping} (addition vs.\ subtraction). 
Tab.~\ref{tab:matchstick_results_textprompt} and Tab.~\ref{tab:matchstick_results_noprompt} summarize results under the text-prompted and pure-visual regimes, respectively.

\paragraph{Move complexity.} 
Two-stick puzzles are consistently more difficult than one-stick puzzles across all models. 
The performance gap is particularly striking in weaker closed-source models, where accuracy often drops below 20\%. 
This reflects the challenge of planning and executing composite edits, which requires maintaining consistency across multiple segments. 
In contrast, one-stick puzzles involve simpler local edits and thus remain more tractable. 
Mixed cases (one/two-move) typically yield higher scores, since many can be solved with an easier one-stick transformation even when a two-stick option exists.

\paragraph{Solution multiplicity.} 
Problems admitting multiple valid corrections prove substantially harder. 
Even strong models such as GPT-o3 and Gemini 2.5 Pro exhibit drops of 20--30 points compared with single-solution cases. 
This indicates that models are highly sensitive to ambiguity in the solution space: 
when several structurally distinct transformations are possible, they often fail to converge on a correct candidate. 
By contrast, single-solution instances provide a unique correction target, 
which better aligns with the deterministic nature of autoregressive decoding.

\paragraph{Operator flipping.} 
Tasks requiring operator changes (\(+\leftrightarrow-\)) constitute a major bottleneck. 
Performance is consistently lower than in no-flip cases, 
suggesting that models are biased toward digit-level manipulations rather than considering operator edits. 
This weakness is particularly pronounced in weaker closed models and universal across all open-source baselines, 
which almost entirely fail on operator-flip puzzles. 
These results highlight the difficulty of extending generalization from digit morphology to abstract symbolic transformations.

\paragraph{Cross-regime comparison.} 
The shift from text-prompted to pure-visual inputs introduces systematic degradation. 
For example, Gemini 2.5 Pro drops from 51.7\% on one-move text-prompted puzzles to 25.8\% in the pure-visual setting. 
This gap underscores OCR and structural parsing as an additional error source, independent of symbolic reasoning itself. 
GPT-o3, while still affected, demonstrates relatively smaller gaps, suggesting stronger robustness to visual noise and layout variability compared with other models.

\paragraph{Open-source models.} 
All tested open-source models (Qwen2.5-VL and InternVL3 families) fail almost entirely across categories, 
with accuracies close to zero. 
This finding emphasizes the current limitations of open-source VLMs in handling structured visual-symbolic reasoning tasks. 
Notably, increasing model scale does not alleviate the problem: 
even the largest variants with 70B+ parameters remain at chance-level performance. 
This suggests that the failure stems not from insufficient capacity, 
but from the lack of targeted supervision and inductive biases for visual–symbolic reasoning. 
It further points to the need for specialized training data and architectural innovations 
capable of bridging continuous perception with discrete symbolic manipulation.

\begin{table*}[!t]
    \scriptsize
    \centering
    \caption{\textbf{Results on the \textsc{MathSticks} Benchmark (w/o text prompt).}
    Accuracy (\%) is reported across categories reflecting move complexity, solution multiplicity, and operator change.
    The best performance is in \textbf{bold}, the second best is \underline{underlined}.}
    \vspace{0.5em}
    \scalebox{1.02}{
    \begin{tabular}{l|ccc|cc|cc}
        \toprule
        \multirow{2}{*}{\textbf{Model}} 
        & \multicolumn{3}{c|}{\textbf{Move Complexity}} 
        & \multicolumn{2}{c|}{\textbf{Solution Multiplicity}} 
        & \multicolumn{2}{c}{\textbf{Operator Change}} \\
        \cmidrule(lr){2-4} \cmidrule(lr){5-6} \cmidrule(lr){7-8}
        & \textbf{1-move} & \textbf{2-move} & \textbf{1/2-move} 
        & \textbf{Single-sol.} & \textbf{Multi-sol.} 
        & \textbf{Sign-changed} & \textbf{No sign-ch.} \\
        \midrule
        \rowcolor[HTML]{F2F2F2} \multicolumn{8}{l}{\textbf{Closed Models}} \\ \midrule
        o3-250416               & \textbf{70.00} & \textbf{32.12} & \textbf{56.70} & \textbf{31.52} & \textbf{45.63} & \textbf{34.20} & \textbf{41.22} \\
        Gemini-2.5-Pro-250506           & \underline{25.83} & \underline{18.22} & \underline{36.63} & \underline{14.76} & \underline{29.72} & \underline{19.70} & \underline{25.05} \\
        Gemini-2.5-Flash-250520         & 18.33 &  5.92 & 15.33 &  4.98 & 11.89 &  8.49 &  8.34 \\
        GPT-o4-mini-250416          & 10.83 &  7.23 & 22.16 &  5.38 & 13.92 & 10.54 &  9.88 \\
        Seed-1.6-Thinking-250615        &  0.00 &  0.68 &  1.85 &  0.00 &  1.55 &  1.61 &  0.58 \\
        Seed-1.6-250615                 &  0.00 &  0.39 &  0.93 &  0.00 &  0.75 &  0.44 &  0.58 \\
        Claude Sonnet 4                 &  0.00 &  0.00 &  0.00 &  0.00 &  0.00 &  0.00 &  0.00 \\
        GPT-4o-241120               &  0.00 &  0.00 &  0.00 &  0.00 &  0.00 &  0.00 &  0.00 \\
        \midrule
        \rowcolor[HTML]{F2F2F2} \multicolumn{8}{l}{\textbf{Open-Source Models}} \\ \midrule
        Qwen2.5-VL-7B-Instruct          & 0.00 & 0.00 & 0.00 & 0.00 & 0.00 & 0.00 & 0.00 \\
        Qwen2.5-VL-32B-Instruct         & 0.00 & 0.00 & 0.00 & 0.00 & 0.00 & 0.00 & 0.00 \\
        Qwen2.5-VL-72B-Instruct         & 0.00 & 0.00 & 0.00 & 0.00 & 0.00 & 0.00 & 0.00 \\
        InternVL3-8B                    & 0.00 & 0.00 & 0.00 & 0.00 & 0.00 & 0.00 & 0.00 \\
        InternVL3-38B                   & 0.00 & 0.00 & 0.00 & 0.00 & 0.00 & 0.00 & 0.00 \\
        InternVL3-78B                   & 0.00 & 0.00 & 0.00 & 0.00 & 0.00 & 0.00 & 0.00 \\
        \bottomrule
    \end{tabular}
    }
    \label{tab:matchstick_results_noprompt}
\vspace{-1.0em}
\end{table*}

\begin{table}[t]
    \centering
    \caption{\textbf{Human performance on the \textsc{MathSticks} benchmark (w/o text prompt).} 
    Accuracy (\%) is reported across difficulty levels together with the average solving time per participant. 
    Since human recognition of digits/operators is error-free, the same accuracies apply under the text-prompted regime.}
    \vspace{0.5em}
    \begin{tabular}{l|ccccc|c}
        \toprule
        \multirow{2}{*}{\textbf{Participant}} & \multicolumn{5}{c|}{\textbf{Accuracy (\%)}} & \multirow{2}{*}{\textbf{Avg. Time}} \\
        \cmidrule(lr){2-6}
         & L1 & L2 & L3 & L4 & Avg. & \\
        \midrule
        P1 & 95.0 & 100.0 & 95.0 & 100.0 & 97.5 & 1m14s \\
        P2 & 100.0 & 95.0 & 72.5 & 73.7 & 85.3 & 1m17s \\
        P3 & 90.0 & 100.0 & 90.0 & 89.5 & 92.4 & 1m06s \\
        \midrule
        \rowcolor[HTML]{F2F2F2} \textbf{Mean} & 95.0 & 98.3 & 85.8 & 87.7 & 91.7 & --- \\
        \rowcolor[HTML]{F2F2F2} \textbf{Std.} & 5.0 & 2.9 & 11.8 & 13.3 & 6.1 & --- \\
        \bottomrule
    \end{tabular}
    \label{tab:human_results}
\end{table}

\subsection{Human Evaluation}
\label{app:human_eval}

To complement model-based evaluation, we conducted a small-scale human study on the \textsc{MathSticks} benchmark. 
Three adult participants (all male, aged 24--30, with university-level education backgrounds) independently solved the full evaluation set under the \textbf{pure-visual regime}, where only the rendered matchstick equations were provided. 
Because the digits and operators are visually unambiguous, participants transcribed the equations without error; hence, the same results apply to the text-prompted regime as well. 

Tab.~\ref{tab:human_results} summarizes participant-level accuracies across difficulty levels together with average solving times. 
On average, humans achieved accuracies above 90\% while spending roughly one minute per problem. 
These results confirm that the benchmark is reliably solvable by humans, yet remains highly challenging for current vision–language models.

\begin{figure}[!t]
    \centering
    \includegraphics[width=\linewidth]{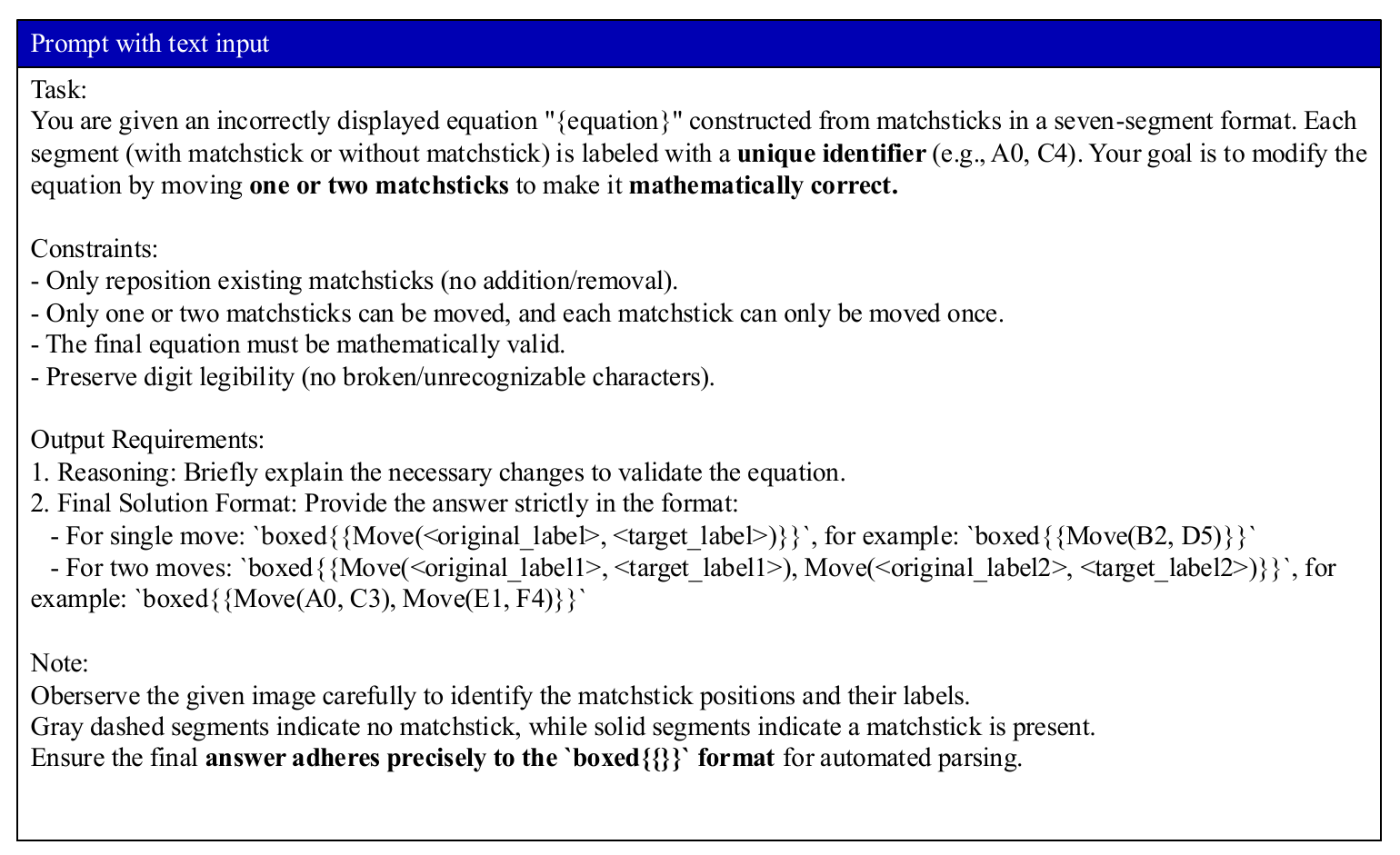}
    \caption{\textbf{Prompt with text input.} The symbolic equation string is provided together with the matchstick rendering.}
    \label{fig:prompt_text}
\end{figure}

\vspace{2em}

\begin{figure}[!t]
    \centering
    \includegraphics[width=\linewidth]{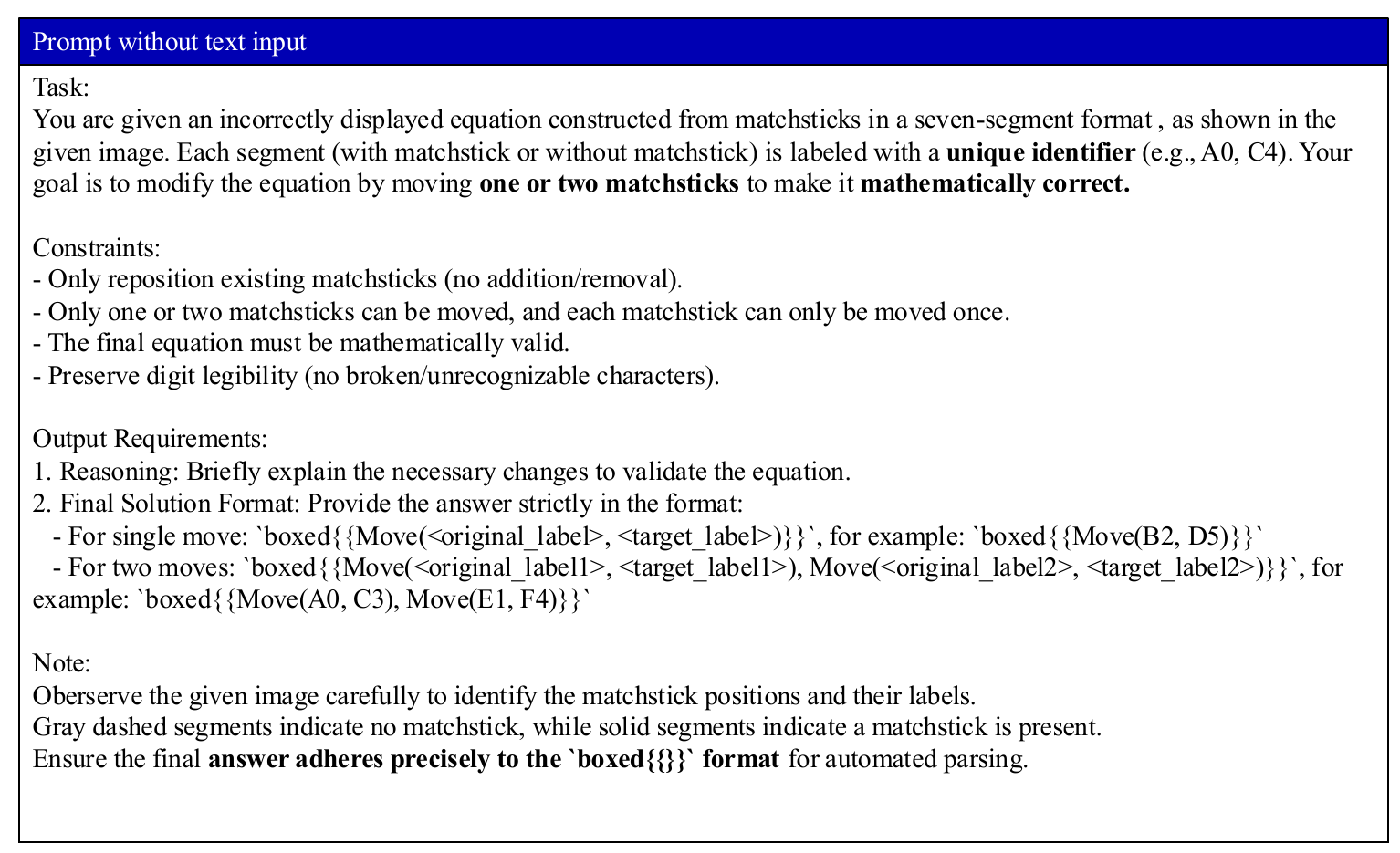}
    \caption{\textbf{Prompt without text input.} Only the matchstick rendering is provided; models must recognize symbols before reasoning.}
    \label{fig:prompt_visual}
\end{figure}

\section{Prompts}
\label{app:prompts}

To ensure reproducibility, we provide the exact prompts used in our evaluation. 
Two regimes are supported: (\textit{i}) \textbf{text-prompted}, where the symbolic equation string is explicitly given; and 
(\textit{ii}) \textbf{pure-visual}, where only the rendered matchstick image is shown, requiring OCR and structural parsing.

\section{Case Studies}
\label{app:case_studies}

\begin{figure}[t]
    \centering
    \includegraphics[width=\linewidth]{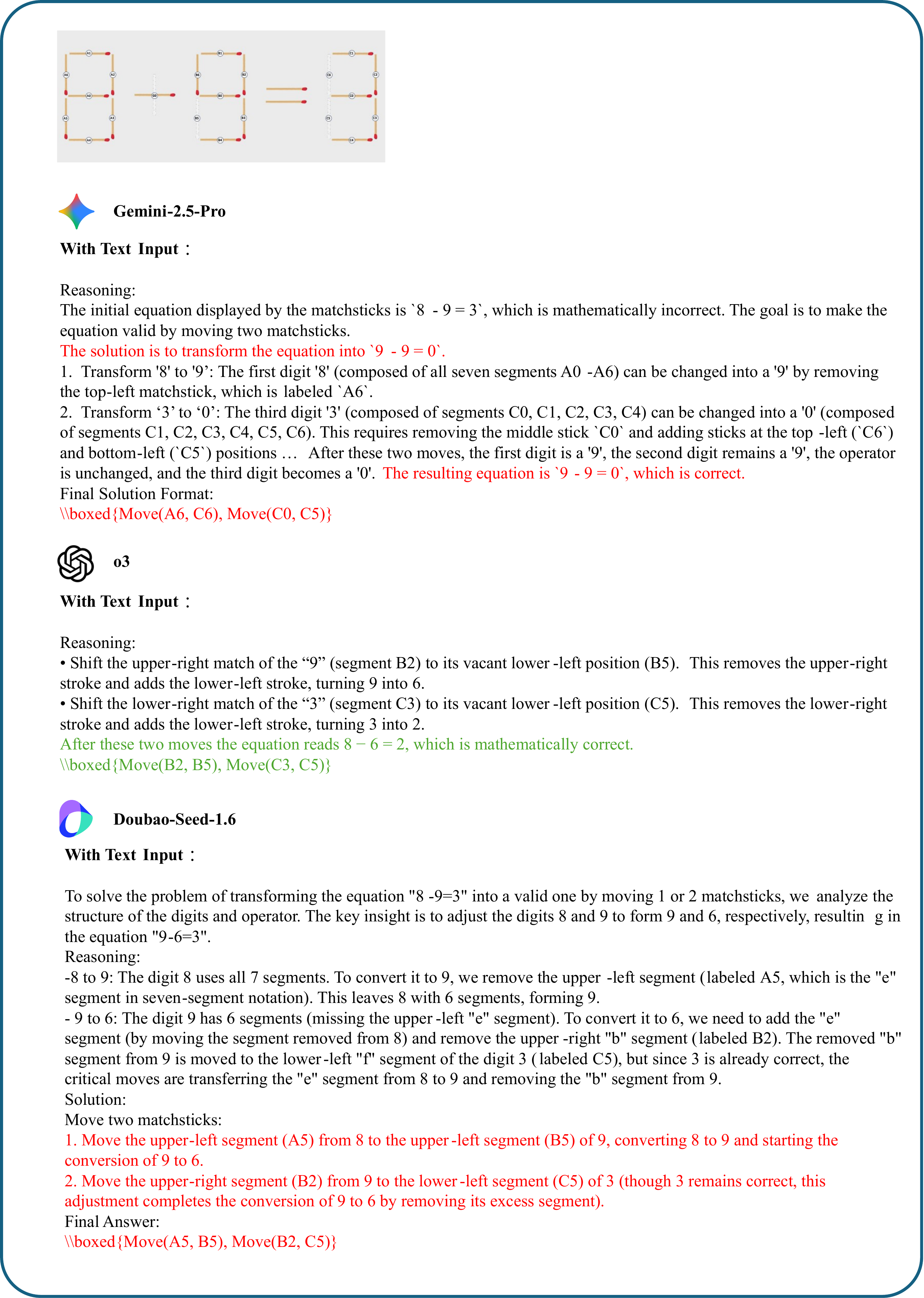}
    \caption{{Case study (Level~1, text-prompted, Case A).}}
    \label{fig:l11}
\end{figure}

\begin{figure}[t]
    \centering
    \includegraphics[width=\linewidth]{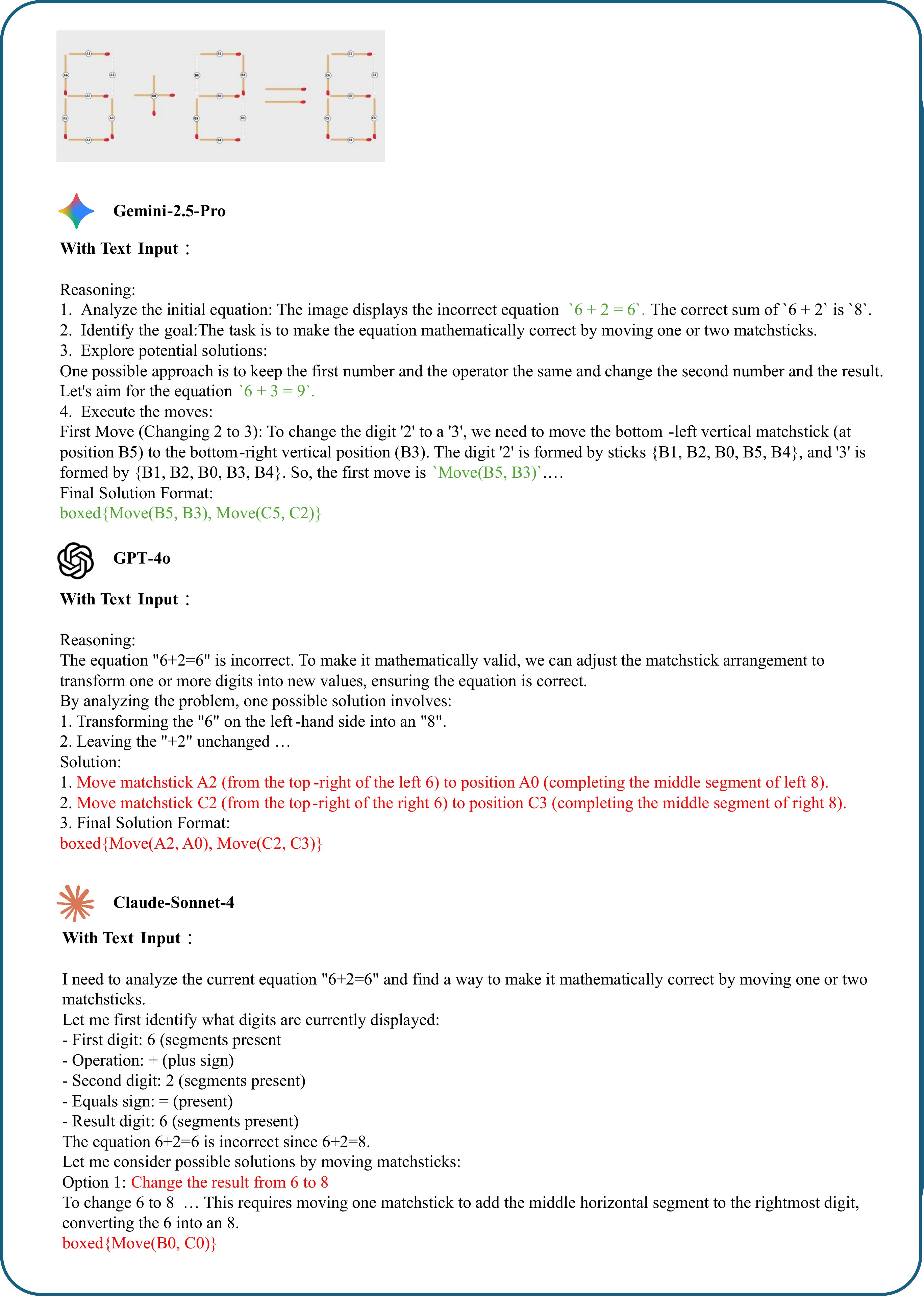}
    \caption{{Case study (Level~1, text-prompted, Case B).} }
    \label{fig:l12}
\end{figure}

\begin{figure}[t]
    \centering
    \includegraphics[width=\linewidth]{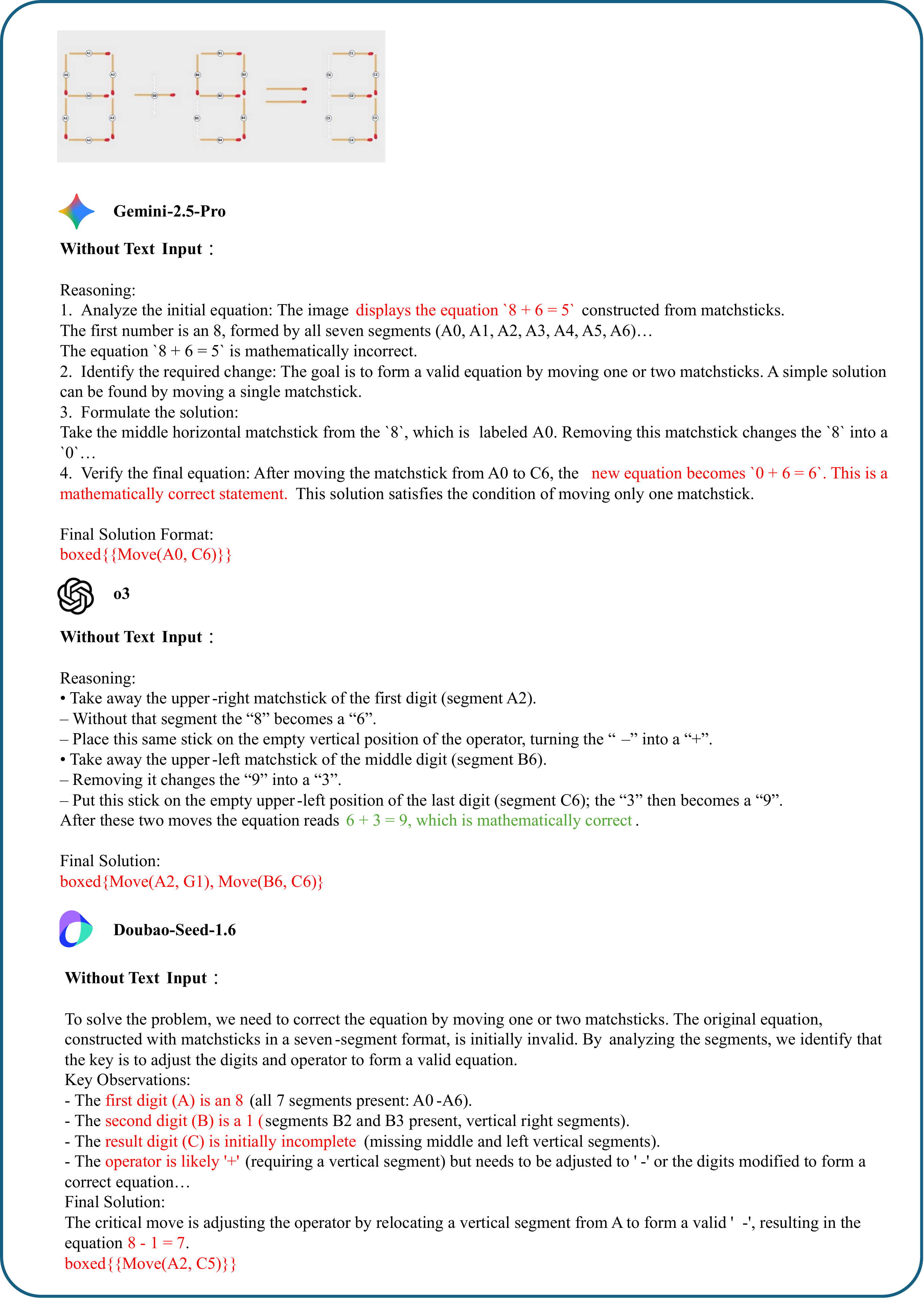}
    \caption{{Case study (Level~1, pure-visual, Case A).} }
    \label{fig:l13}
\end{figure}

\begin{figure}[t]
    \centering
    \includegraphics[width=\linewidth]{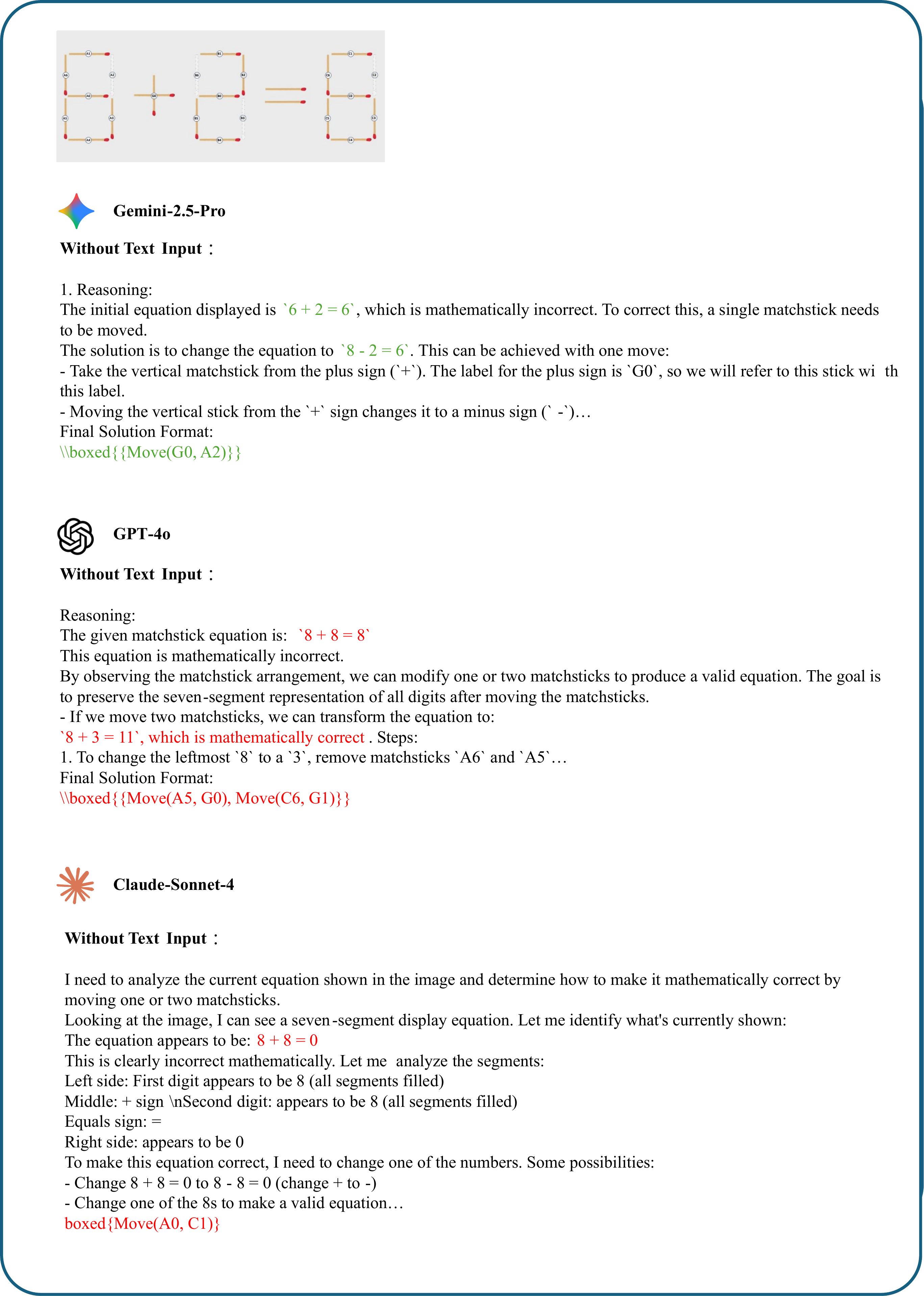}
    \caption{{Case study (Level~1, pure-visual, Case B).} }
    \label{fig:l14}
\end{figure}

\begin{figure}[t]
    \centering
    \includegraphics[width=\linewidth]{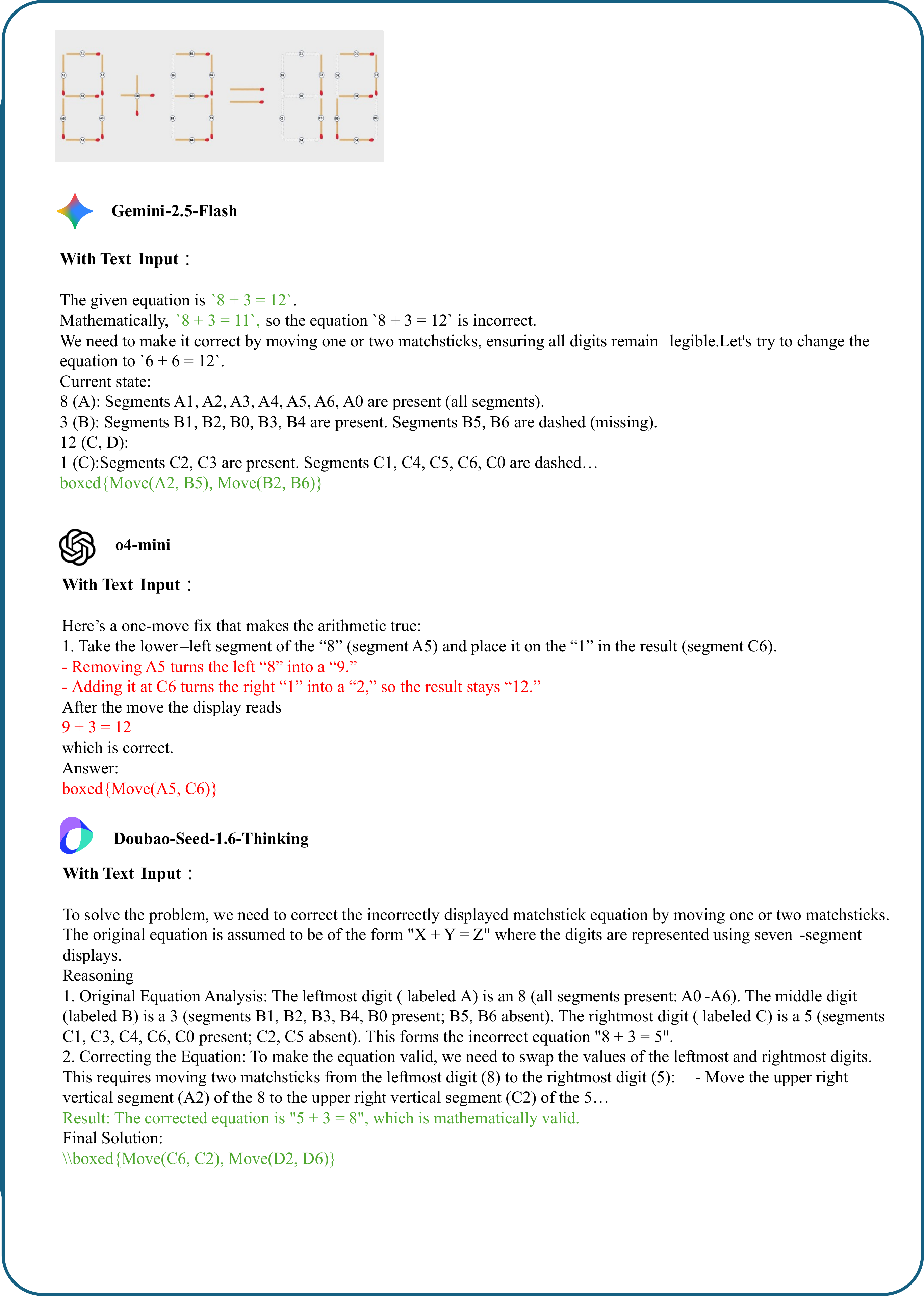}
    \caption{{Case study (Level~2, text-prompted, Case A).} }
    \label{fig:l21}
\end{figure}

\begin{figure}[t]
    \centering
    \includegraphics[width=\linewidth]{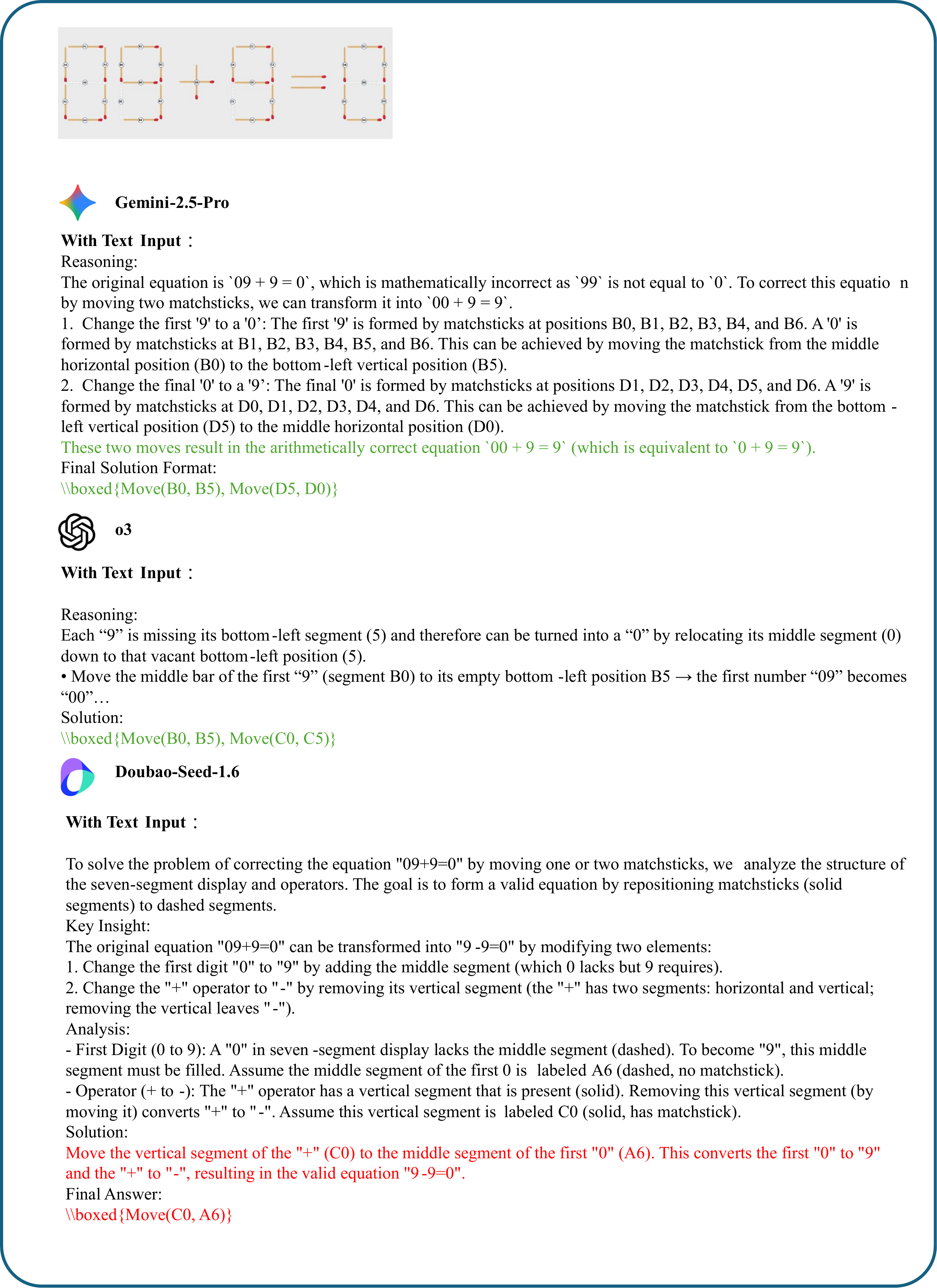}
    \caption{{Case study (Level~2, text-prompted, Case B).} }
    \label{fig:l22}
\end{figure}

\begin{figure}[t]
    \centering
    \includegraphics[width=\linewidth]{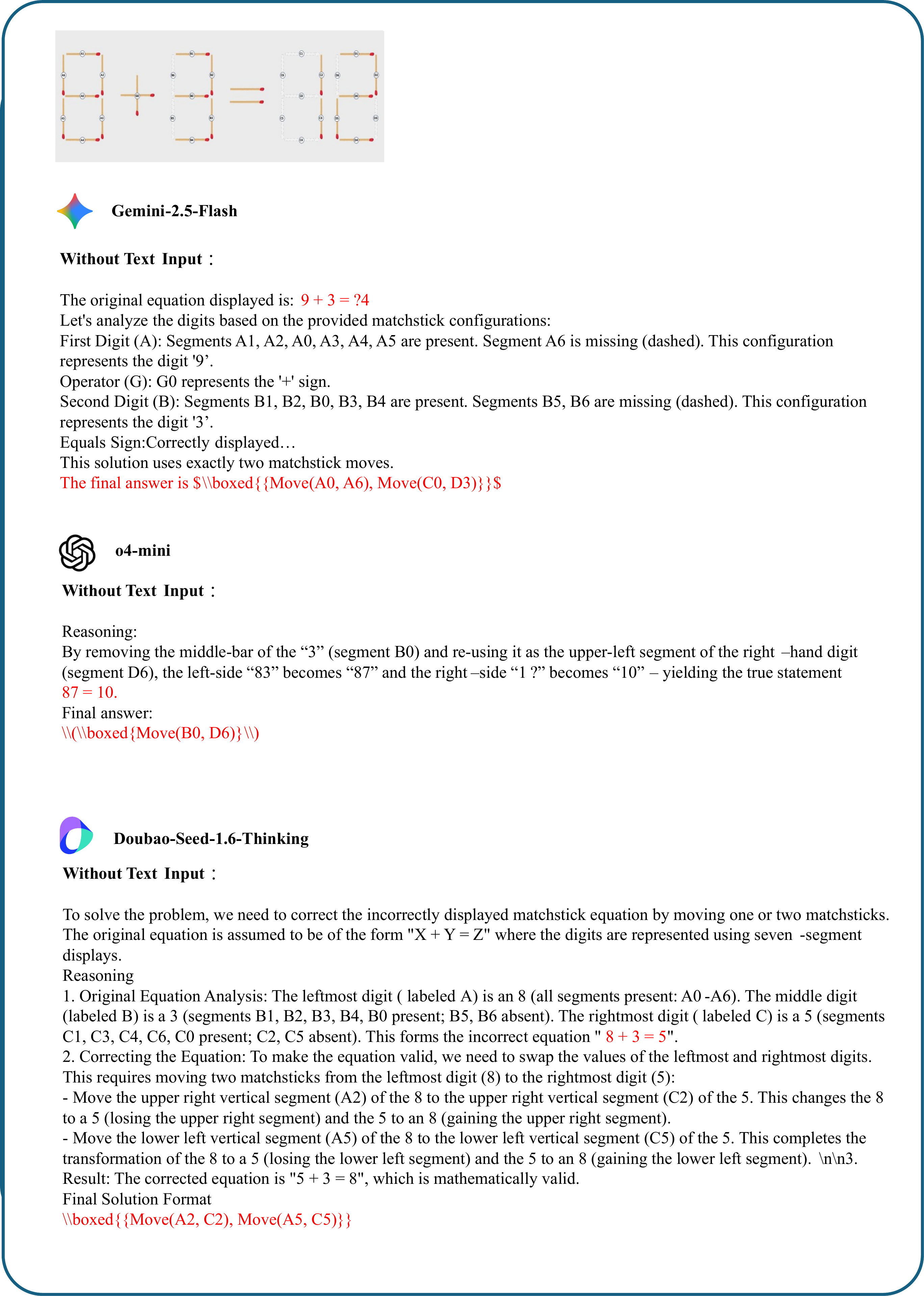}
    \caption{{Case study (Level~2, pure-visual, Case A).} }
    \label{fig:l23}
\end{figure}

\begin{figure}[t]
    \centering
    \includegraphics[width=\linewidth]{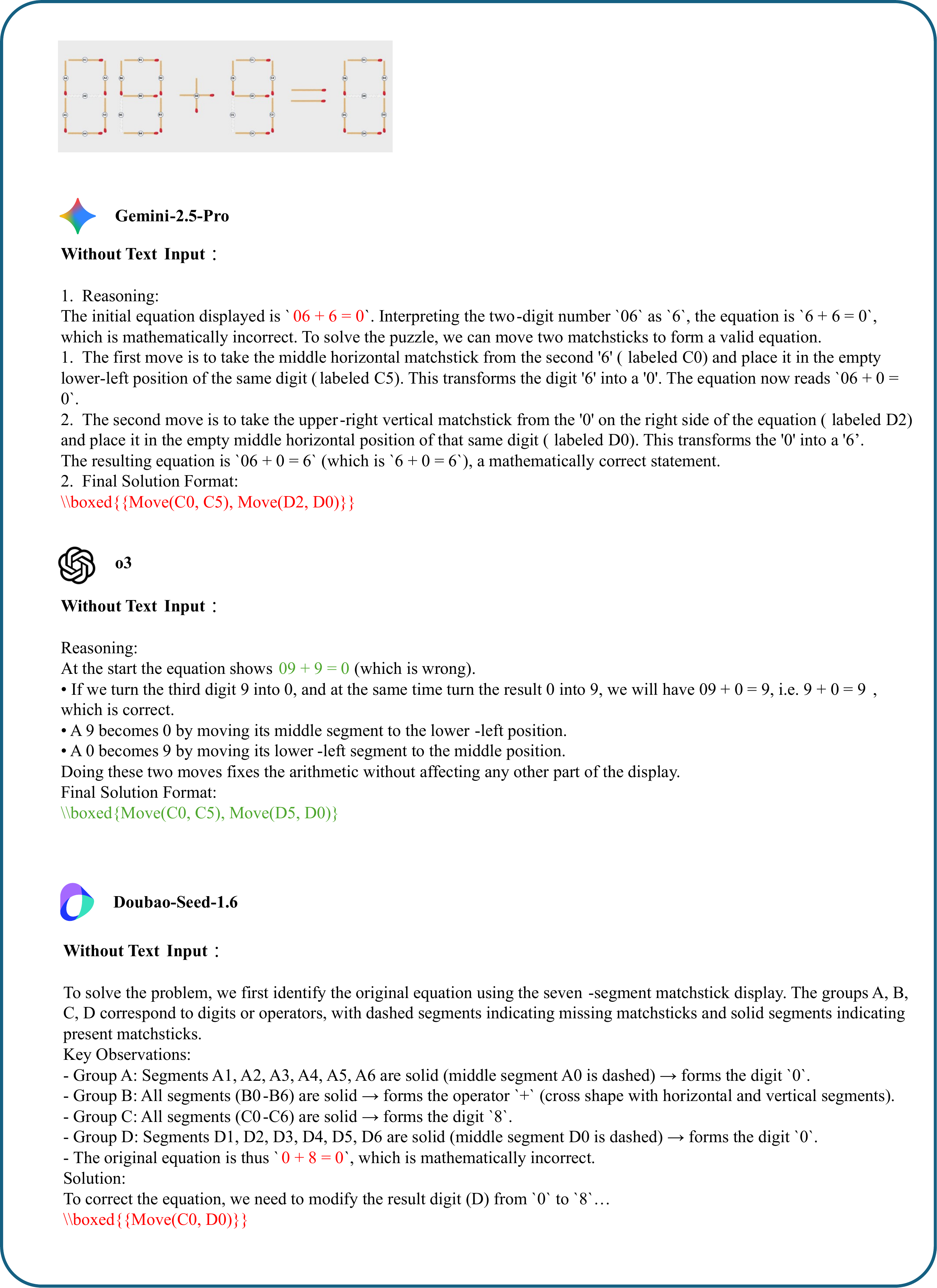}
    \caption{{Case study (Level~2, pure-visual, Case B).} }
    \label{fig:l24}
\end{figure}

\begin{figure}[t]
    \centering
    \includegraphics[width=\linewidth]{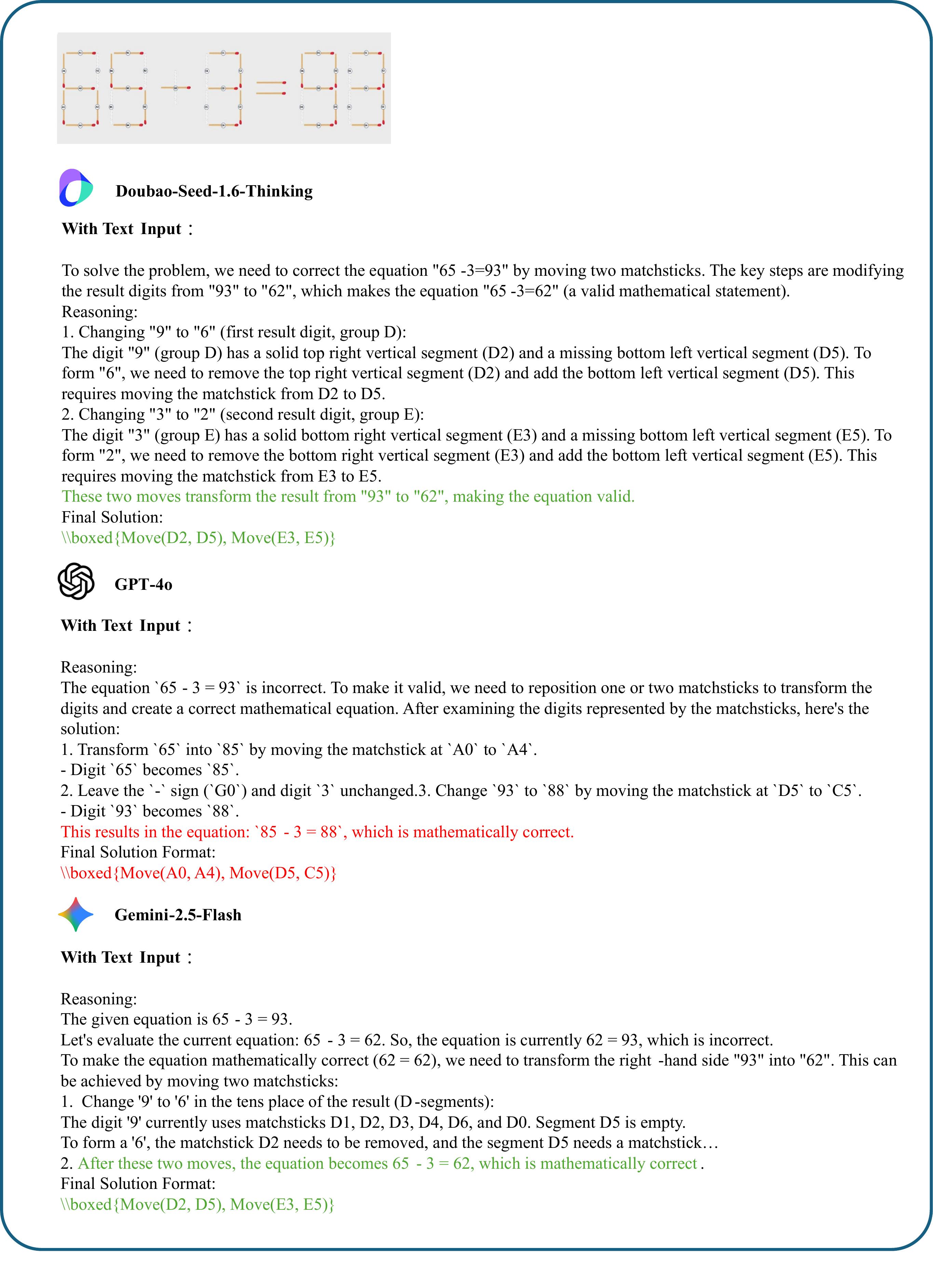}
    \caption{{Case study (Level~3, text-prompted, Case A).} }
    \label{fig:l31}
\end{figure}

\begin{figure}[t]
    \centering
    \includegraphics[width=\linewidth]{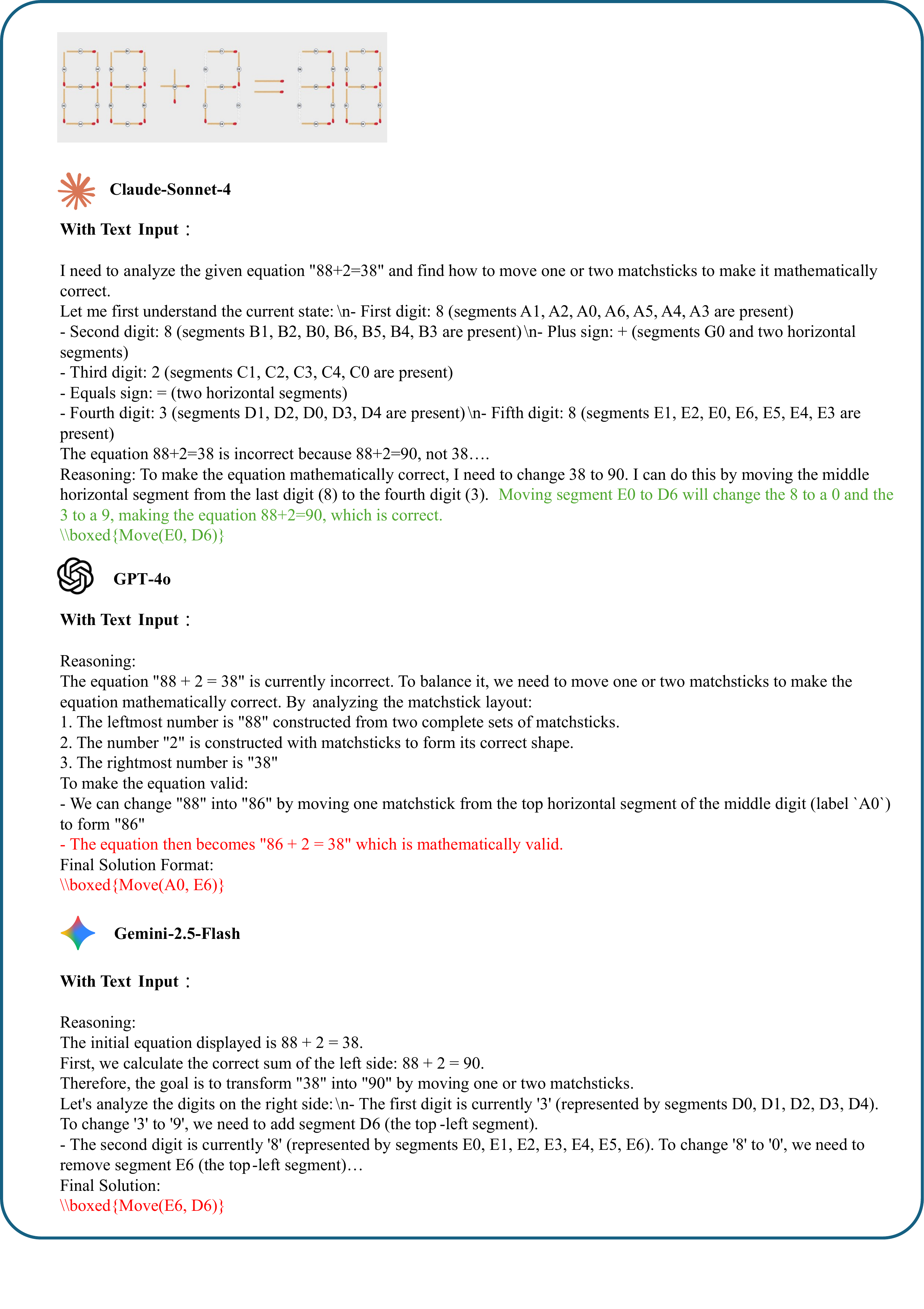}
    \caption{{Case study (Level~3, text-prompted, Case B).} }
    \label{fig:l32}
\end{figure}

\begin{figure}[t]
    \centering
    \includegraphics[width=\linewidth]{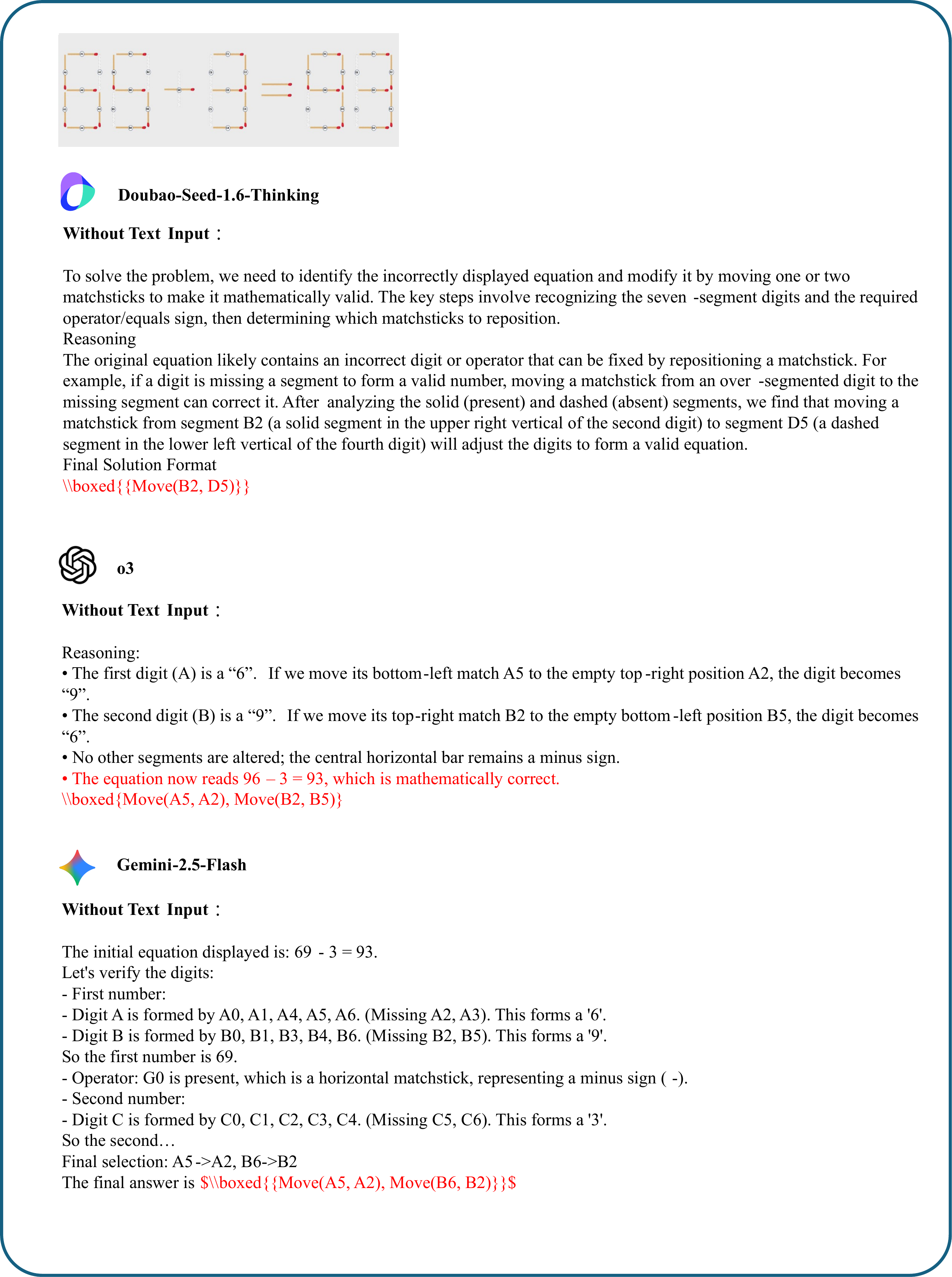}
    \caption{{Case study (Level~3, pure-visual, Case A).} }
    \label{fig:l33}
\end{figure}

\begin{figure}[t]
    \centering
    \includegraphics[width=\linewidth]{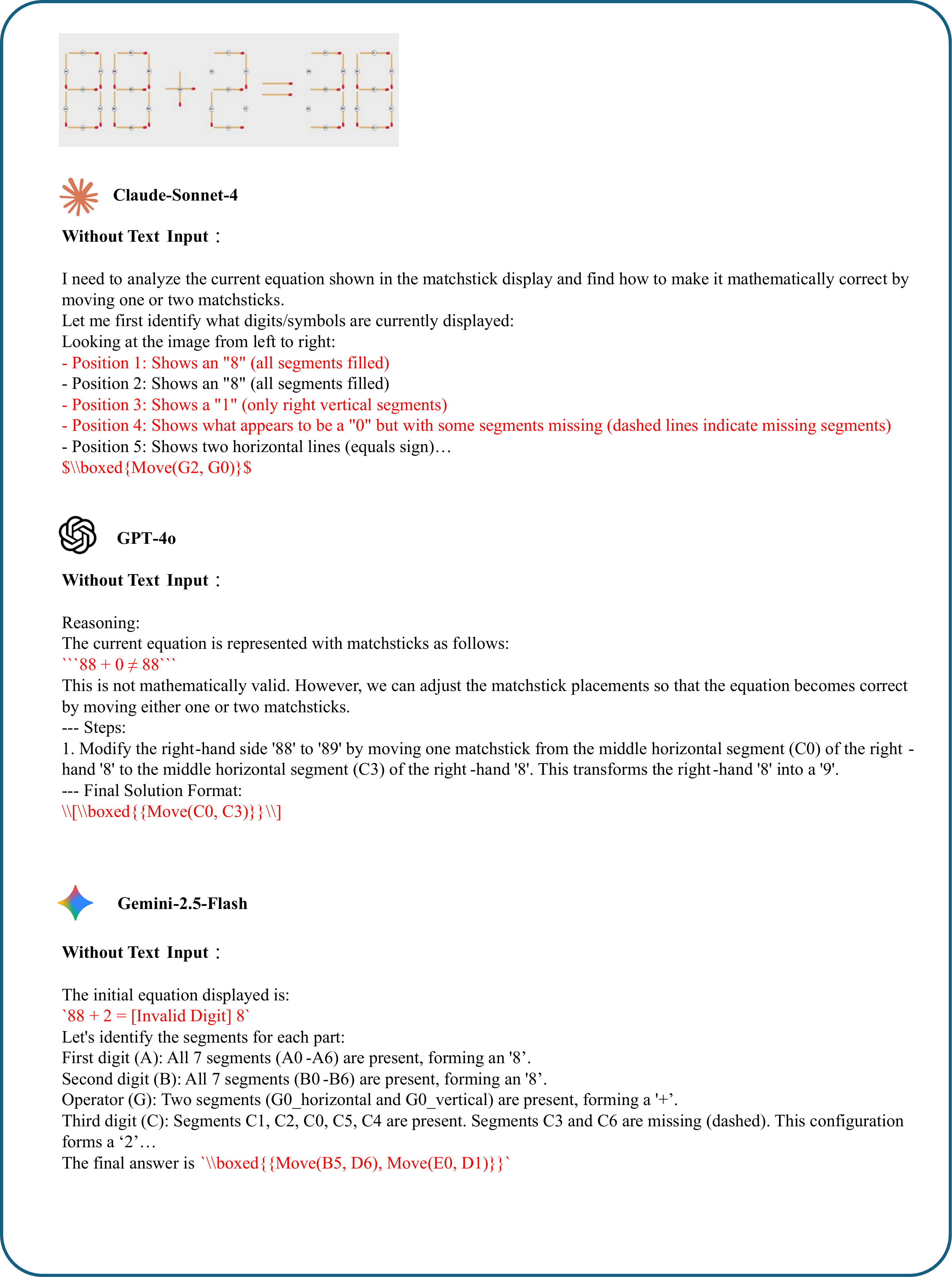}
    \caption{{Case study (Level~3, pure-visual, Case B).} }
    \label{fig:l34}
\end{figure}

\begin{figure}[t]
    \centering
    \includegraphics[width=\linewidth]{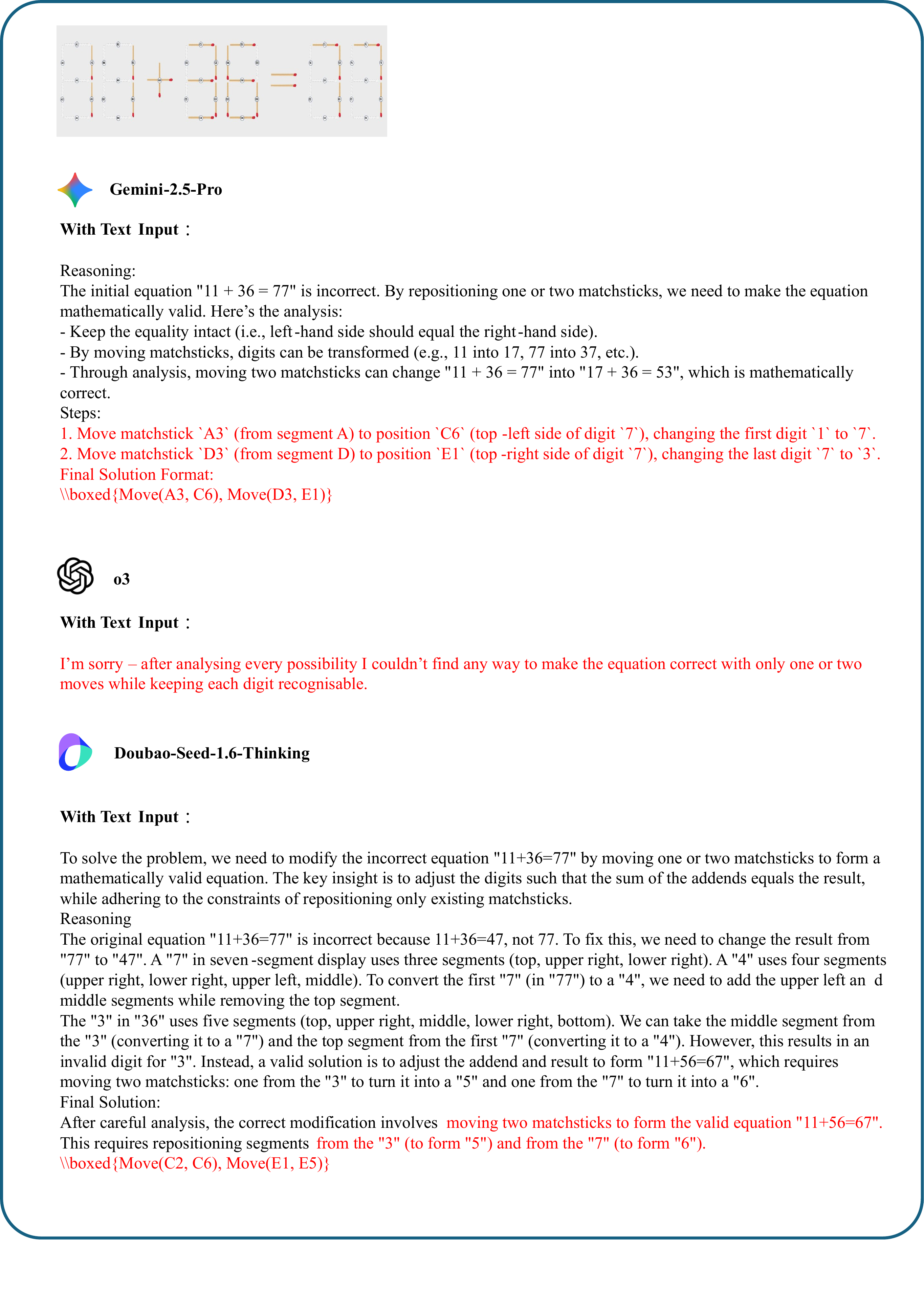}
    \caption{{Case study (Level~4, text-prompted, Case A).} }
    \label{fig:l41}
\end{figure}

\begin{figure}[t]
    \centering
    \includegraphics[width=\linewidth]{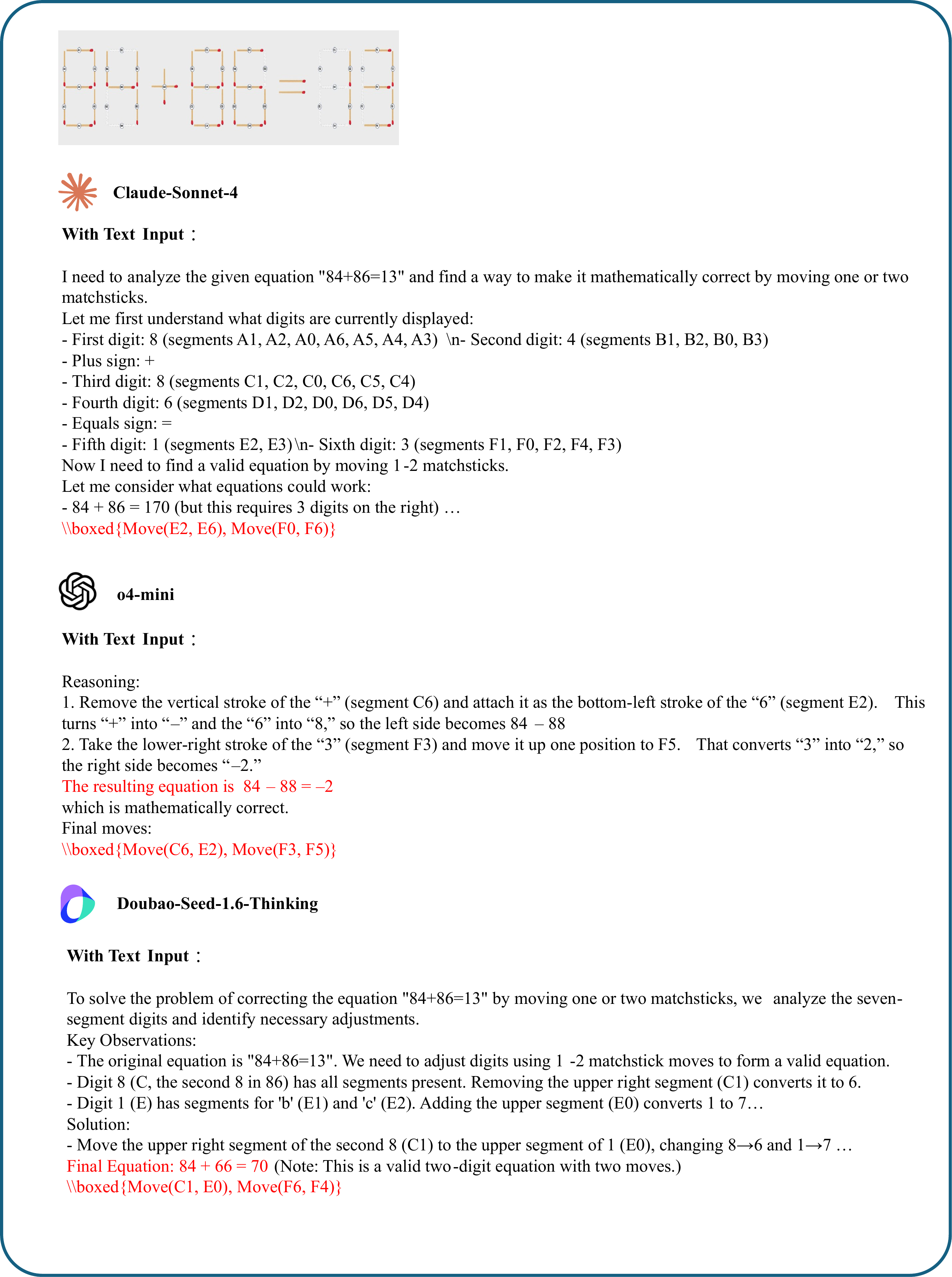}
    \caption{{Case study (Level~4, text-prompted, Case B).} }
    \label{fig:l42}
\end{figure}

\begin{figure}[t]
    \centering
    \includegraphics[width=\linewidth]{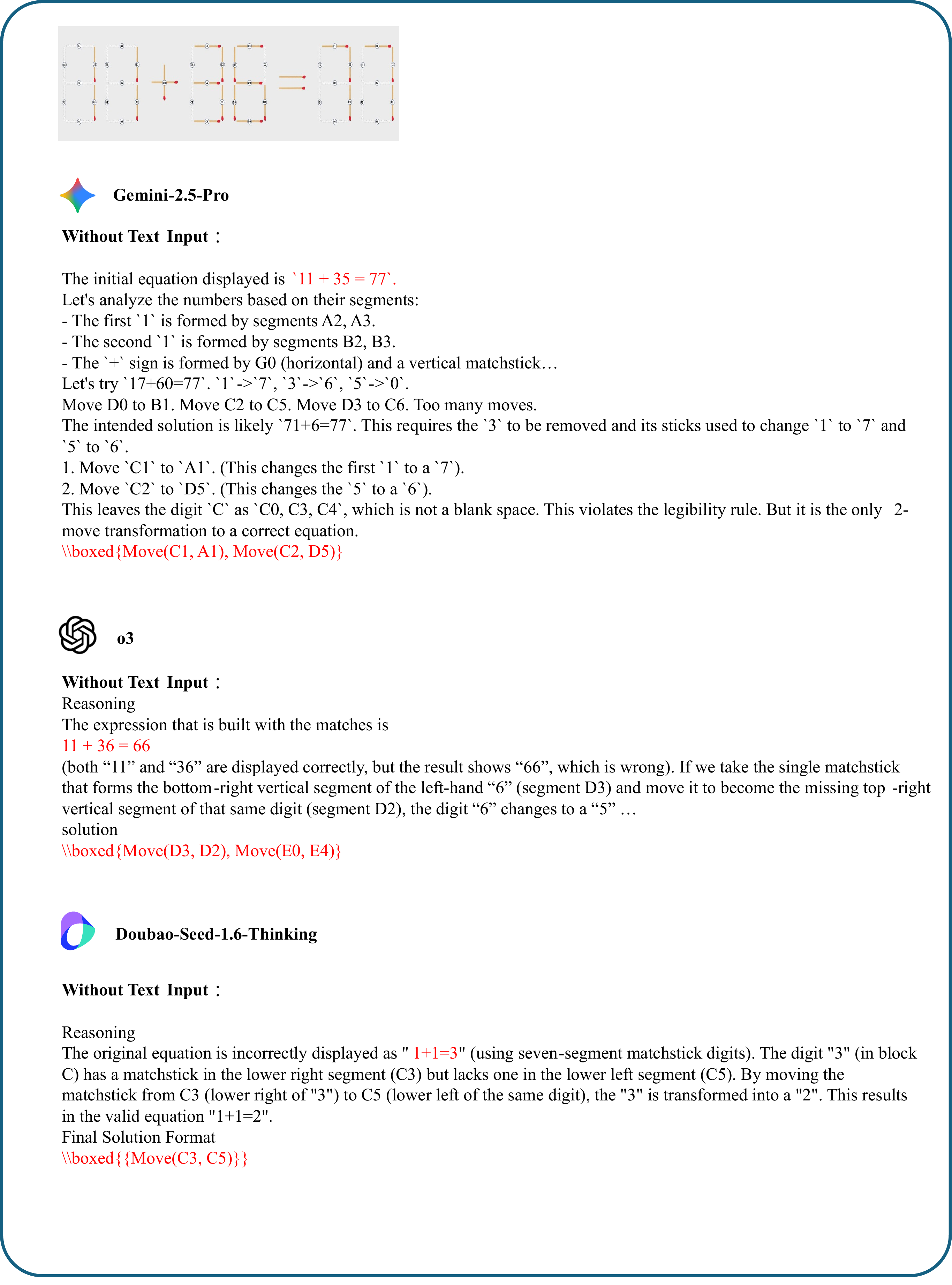}
    \caption{{Case study (Level~4, pure-visual, Case A).} }
    \label{fig:l43}
\end{figure}

\begin{figure}[t]
    \centering
    \includegraphics[width=\linewidth]{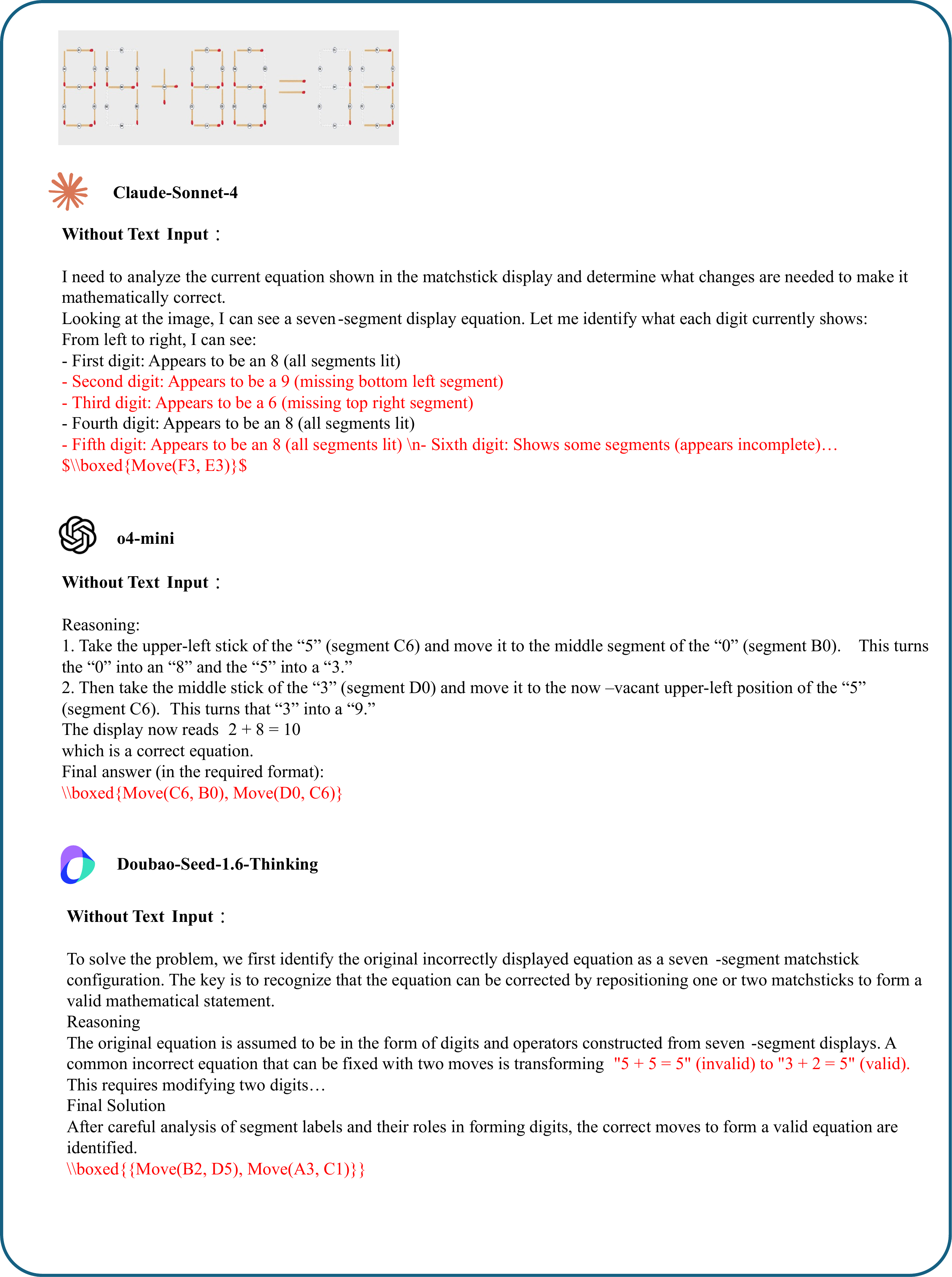}
    \caption{{Case study (Level~4, pure-visual, Case B).} }
    \label{fig:l44}
\end{figure}

To qualitatively illustrate model behavior, we sampled two representative puzzles from each difficulty level (L1–L4), under both input regimes (text-prompted and pure-visual), resulting in 16 cases in total. 
For each case, we present the input matchstick puzzle (i.e., the incorrect equation to be solved) and compare predictions from multiple representative models. Correct solutions are not explicitly annotated in the figure but can be verified against the solvability constraints of the benchmark.
This analysis highlights common failure modes, including: incorrect operator flipping, partial edits that leave digits illegible, and failure to converge when multiple solutions exist. 
Fig.~\ref{fig:l11}–Fig.~\ref{fig:l44} present the visualizations.

\end{document}